\documentclass[lettersize,journal]{IEEEtran}
\usepackage{amsmath,amsfonts}
\usepackage{algorithmic}
\usepackage{algorithm}
\usepackage{array}
\usepackage{textcomp}
\usepackage{stfloats}
\usepackage{url}
\usepackage{verbatim}
\usepackage{graphicx}
\usepackage{cite}

\usepackage{amssymb}
\usepackage{booktabs}
\usepackage{adjustbox}
\usepackage{multirow}
\usepackage{pifont}
\usepackage{color}
\usepackage{subfigure}
\usepackage{url}
\usepackage{amsthm}
\usepackage{amsfonts}
\usepackage{caption}
\usepackage{threeparttable}
\usepackage{enumitem}

\usepackage[capitalize]{cleveref}
\crefname{section}{Sec.}{Secs.}
\Crefname{section}{Section}{Sections}
\Crefname{table}{Table}{Tables}
\crefname{table}{Tab.}{Tabs.}

\newcommand{\whiteding}[1]{\ding{\numexpr171+#1\relax}}

\begin{document}

\title{Hybrid Tracker with Pixel and Instance for Video Panoptic Segmentation}

\author{Weicai Ye*, Xinyue Lan*, Ge Su, Hujun Bao, Zhaopeng Cui and Guofeng Zhang

\thanks{W. Ye, X. Lan, H. Bao, Z. Cui and G. Zhang are with the State Key Lab of CAD\&CG, Zhejiang University, and affiliated with ZJU-SenseTime Joint Lab of 3D Vision. G. Su is with Zhejiang University. E-mails: \{weicaiye, xinyuelan, suge, baohujun, zhpcui, zhangguofeng\}@zju.edu.cn}
\thanks{*: indicates equal contribution. G. Zhang is the corresponding author.}
}





\maketitle

\begin{abstract}

Video Panoptic Segmentation (VPS) aims to generate coherent panoptic segmentation and track the identities of all pixels across video frames. Existing methods predominantly utilize the trained instance embedding to keep the consistency of panoptic segmentation. However, they inevitably struggle to cope with the challenges of small objects, similar appearance but inconsistent identities, occlusion, and strong instance contour deformations. To address these problems, we present HybridTracker, a lightweight and joint tracking model attempting to eliminate the limitations of the single tracker. HybridTracker performs pixel tracker and instance tracker in parallel to obtain the association matrices, which are fused into a matching matrix. In the instance tracker, we design a differentiable matching layer, ensuring the stability of inter-frame matching. In the pixel tracker, we compute the dice coefficient of the same instance of different frames given the estimated optical flow, forming the Intersection Over Union (IoU) matrix. We additionally propose mutual check and temporal consistency constraints during inference to settle the occlusion and contour deformation challenges. Comprehensive experiments show that HybridTracker achieves superior performance than state-of-the-art methods on Cityscapes-VPS and VIPER datasets. 

\end{abstract}


\section{Introduction}
\label{sec:intro}
\begin{figure}
\centering
\includegraphics[width=\linewidth]{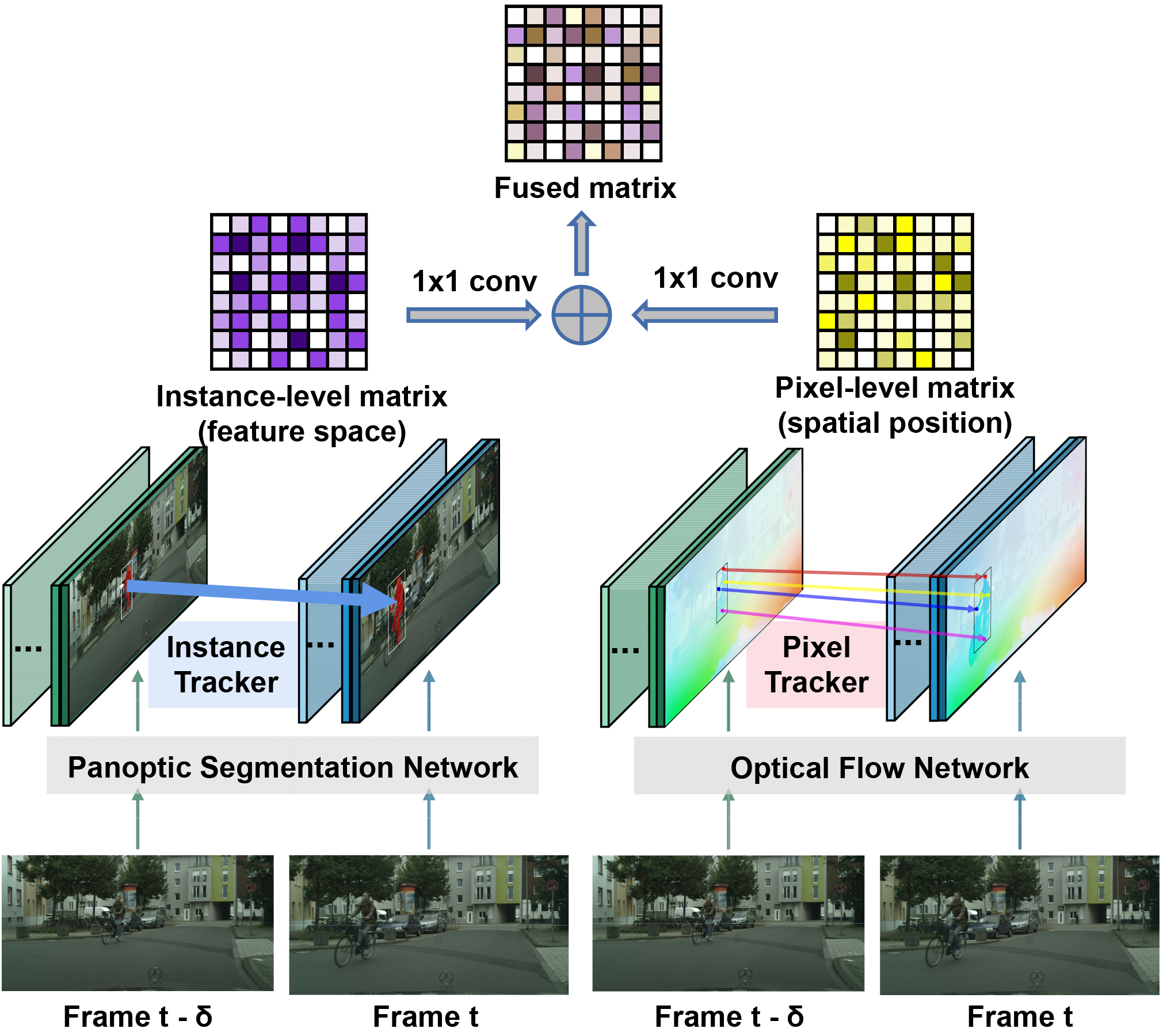}
\caption{\textbf{Brief illustration of HybridTracker.} HybridTracker achieves video panoptic segmentation from two perspectives: feature space (Instance Tracker) and spatial location (Pixel Tracker).
}
\label{fig:teaser Hybrid Tracker}
\end{figure}

\IEEEPARstart{V}{ideo} panoptic segmentation (VPS) aims to assign semantic classes and generate consistent association of instance IDs across video frames with many applications ranging from scene understanding in autonomous driving, video surveillance in smart cities, video editing in entertainment manipulation, etc. 
Existing methods tackling the problem generally perform the image panoptic segmentation and obtain the matching relationship between adjacent images by calculating the similarity of the instance embedding.

A pioneer work, proposed by~\cite{kim2020vps}, first formally defined a brand new task of video panoptic segmentation with a new re-organized dataset and presented a network, i.e., VPSNet, to solve this problem. 
VPSNet extends UPSNet~\cite{xiong2019upsnet} and adds the MaskTrack head~\cite{yang2019video} to track each instance, which can track larger objects robustly. However, some small objects are difficult to accurately segment due to the limited image segmentation accuracy (see Fig.~\ref{fig:comparison test vps} \whiteding{6}). It also struggles to learn a discriminative embedding, especially for some objects with similar appearance but inconsistent identities (see Fig.~\ref{fig:comparison val vps} \whiteding{2}), which may lead to ambiguous tracking. Furthermore, VPSNet uses optical flow estimation to fuse the features of adjacent frames to further improve the segmentation, which is undoubtedly heavy and complicated. Woo et al.~\cite{woo2021learning} improved VPSNet by introducing the contrast loss and pixel-level tube-matching loss to obtain time-invariant features,
while it still relies on the instance-level features for the tracking as VPSNet.

\begin{figure*}[t]
  \centering
  \includegraphics[width=\linewidth]{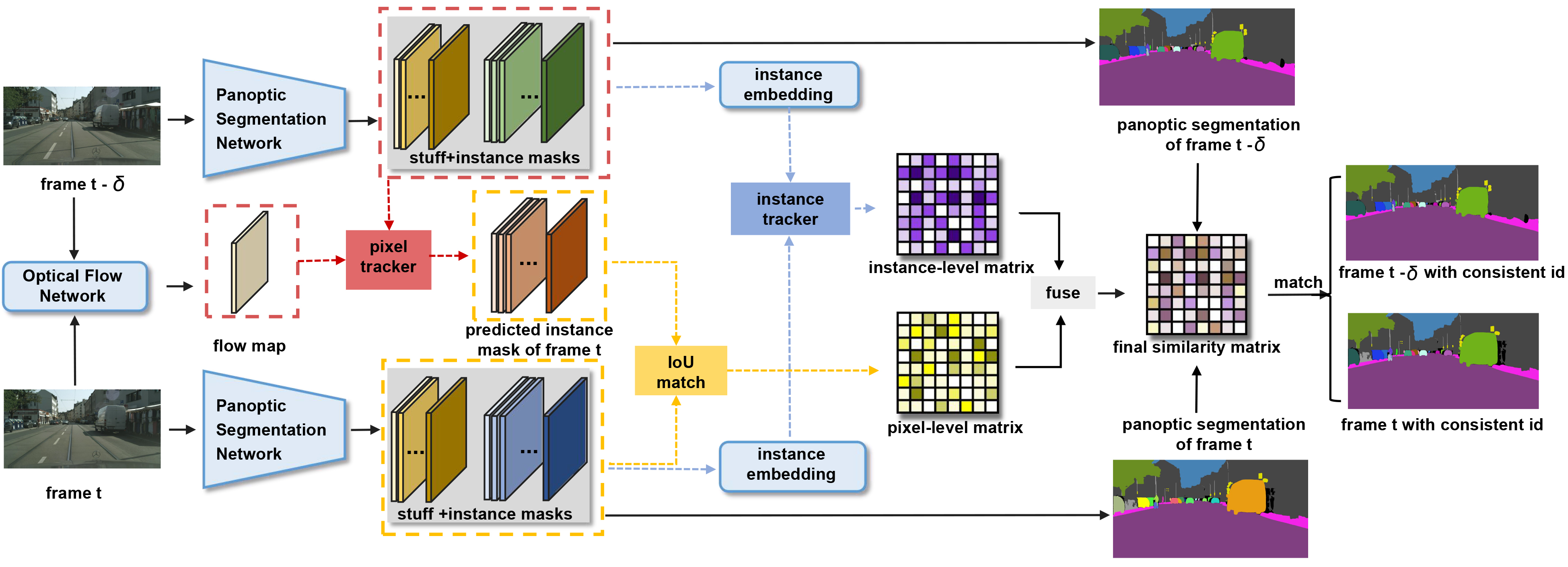}
  \caption{\textbf{HybridTracker framework.} Our network takes consecutive video frames as input and outputs the segmentation results of each frame and the consistent tracking IDs between frames. Each frame is fed into the image panoptic segmentation network. Then instance tracker and pixel tracker are run to obtain the correlation matrices of instances between frames, which are finally fused into a matching matrix to obtain consistent video panoptic segmentation results.}
  \label{fig:method.framework hybrid}
\end{figure*}

CenterTrack~\cite{zhou2020tracking} proposed a simultaneous detection and tracking algorithm to localize objects and predict their associations by objects' centers, which can be considered a tracker based on pixels. However, the tracking is not stable enough due to the sparsity of the objects' center, especially in the face of multiple objects occlusion and reappearance after the disappearance. Moreover, it has not been applied to the video panoptic segmentation task. Existing instance-based and pixel-based trackers are particularly weak in handling multi-object occlusion (see Fig.~\ref{fig:comparison val vps} \whiteding{3} and \ref{fig:comparison test vps} \whiteding{5}) and large instance contour deformations (see Fig.~\ref{fig:comparison ablation study hybridtracker temporal constraints} \whiteding{3}).

To address the problems aforementioned, we propose to track with both pixel and instance for video panoptic segmentation. Our insight is that VPS can benefit from different perspectives including feature space (instance tracker) and spatial location (pixel tracker), which can effectively mitigate the drawbacks of using a single-level tracker (see Fig.~\ref{fig:teaser Hybrid Tracker}). However, directly combining the existing instance-based and pixel-based tracker is non-trivial. On the one hand, it leads to onerous network parameters, making the models much larger and harder to train on a standard graphics card. On the other hand, different trackers need to maintain a good tracking performance to obtain complimentary tracking effects through fusion. Otherwise, intermediate results will be obtained.

To achieve a trade-off between memory consumption and VPS performance, we propose a lightweight and joint tracking network, termed HybridTracker, (see Fig.~\ref{fig:method.framework hybrid}). HybridTracker performs panoptic segmentation for each frame and adopts different trackers to obtain the correlation matrices fused into a final matching matrix to obtain the spatiotemporal consistent video panoptic segmentation results. 
Specifically, we use a lightweight image panoptic segmentation model PanopticFCN~\cite{li2021panopticfcn} to obtain the panoptic segmentation results for each frame, which can obtain competitive segmentation results with fewer parameters and faster speed. 
In the instance tracker, to reduce memory consumption, we take the segmented instance mask to obtain the ROI mask of each instance by cropping, scaling, and padding, and then leverage two fully connected layers to obtain the embedding of each instance.

To address the issue of training convergence that usually exists for the instance-based tracker~\cite{hermans2017defense, Ye2021SuperPlane}, we present a differentiable matching layer to obtain the correspondence between different instances and then train the network with the cross-entropy loss.
In pixel tracker, rather than sparse objects' center in CenterTrack~\cite{zhou2020tracking}, we use the dense optical flow estimation network RAFT~\cite{teed2020raft} to obtain the dense pixel motion of the current frame and add the original coordinate of pixels to obtain the mask of each instance in the next frame. The IoU matching matrix is obtained by computing the dice coefficient of the same instance of different frames. 

To deal well with the irregular deformations caused by perspective projection and individual instances reappearing after disappearing, we also propose mutual check and temporal consistency constraints applied to the proposed HybridTracker, which can effectively eliminate these issues. Our contributions can be briefly summarised as four-fold.

\begin{itemize}
    \item We present a lightweight and efficient hybrid tracker network consisting of an instance tracker and a pixel tracker that effectively copes with video panoptic segmentation.
    \item Our differentiable matching module for instance tracker can smoothly train the embedding of instances and maintain tracking robustness. While our dense pixel tracker does not require additional training.
    \item Our proposed mutual track and temporal consistency constraints can effectively handle complex challenges such as reappearance after a disappearance due to occlusion and strong instance deformations.
    \item Extensive experiments demonstrate that HybridTracker achieves superior performance than state-of-the-art methods on Cityscapes-VPS and VIPER datasets.
\end{itemize}

\section{Related Work}
\label{sec:related work}

\textbf{Video Segmentation.} A comprehensive overview~\cite{wang2021survey} of multiple tasks in video segmentation has recently been proposed, which broadly classifies video segmentation into eight tasks such as video semantic segmentation~\cite{li2021video, hoyer2021three, chen2020naive, liu2020efficient}, interactive video object segmentation~\cite{yin2021learning, heo2021guided, cheng2021modular}, semi-automatic video object segmentation~\cite{xu2020segment, li2020delving, xie2021efficient, duke2021sstvos}, object-level automatic video object segmentation~\cite{ren2021reciprocal, porzi2020learning, luiten2020unovost, yang2021dystab}, language-guided video object segmentation~\cite{hui2021collaborative, ye2021referring, mcintosh2020visual}, instance-level automatic video object segmentation~\cite{zhou2021target, ventura2019rvos}, video instance segmentation~\cite{yang2019video, wang2020end, athar2020stem, bertasius2020classifying, Cao_SipMask_ECCV_2020, ying2021srnet, lin2021video}, and video panoptic segmentation~\cite{kim2020vps, woo2021learning, qiao2021vip, Weber2021NEURIPSDATA}. Among them, it systematically describes the methods used in these tasks in recent years, the datasets used and the results achieved so far, as well as the future trends. In this work, we focus on video panoptic segmentation.

\textbf{Video Panoptic Segmentation.} The pioneering work of video panoptic segmentation -- VPS~\cite{kim2020vps} extends the panoptic segmentation task from the image domain to the video domain, proposing an instance-level tracking-based approach. Based on this, SiamTrack~\cite{woo2021learning} proposes pixel-tube matching loss and contrast loss, which improves the discriminative power of instance embedding further. These methods can be roughly considered as instance trackers. The contemporaneous work, STEP~\cite{Weber2021NEURIPSDATA} proposes to directly segment and track each pixel of the video. In comparison, we propose a hybrid tracker, which consists of a pixel tracker and an instance tracker for video panoptic segmentation. Note that we focus on better tracking of the video segmentation results, rather than fusing the features of different frames to obtain better image segmentation results as in VPSNet~\cite{kim2020vps}. We also add the mutual check and temporal consistency constraints to tackle the challenges of deformations and reappearance after occlusion. VIP-DeepLab~\cite{qiao2021vip} additionally introduces depth information to improve the performance of VPS, while our work focuses only on the video track without considering depth.

\textbf{Video Tracking.} Video segmentation requires consistent segmentation between consecutive frames, which is essentially the tracking of video elements. VPS~\cite{kim2020vps} achieved this by tracking each instance from the perspective of feature space, while CenterTrack~\cite{zhou2020tracking} through the center of each instance from the perspective of spatial location. No doubt that they struggle to handle some complex challenges, such as occlusion. A considerable amount of work~\cite{zhang2020fair, Pang_2021_CVPR, wang2019towards} exists that focuses on multiple object tracking rather than video panoptic segmentation. QDTrack~\cite{Pang_2021_CVPR} proposed quasi-dense samples to track multiple objects, but it's still based on feature space. Our proposed HybridTracker, which consists of an instance tracker and a dense pixel tracker, can robustly cope with the complex scenes for video panoptic segmentation from two perspectives: feature space and spatial location.

\section{Methodology}
Given a monocular video, our target is to generate coherent video panoptic segmentation. To this end, we propose a lightweight and efficient network, termed HybridTracker (see Fig.~\ref{fig:method.framework hybrid}) to tackle this problem. HybridTracker consists of four main components: an image panoptic segmentation network with shared weights, an instance tracker module with a differentiable matching layer, a dense pixel tracker module, and a final fusion module to obtain the matching matrix. To handle some complex challenges, we additionally add a mutual check and temporal consistency constraints to the instance tracker and pixel tracker during inference. 

\begin{figure}
\centering
\includegraphics[width=\linewidth]{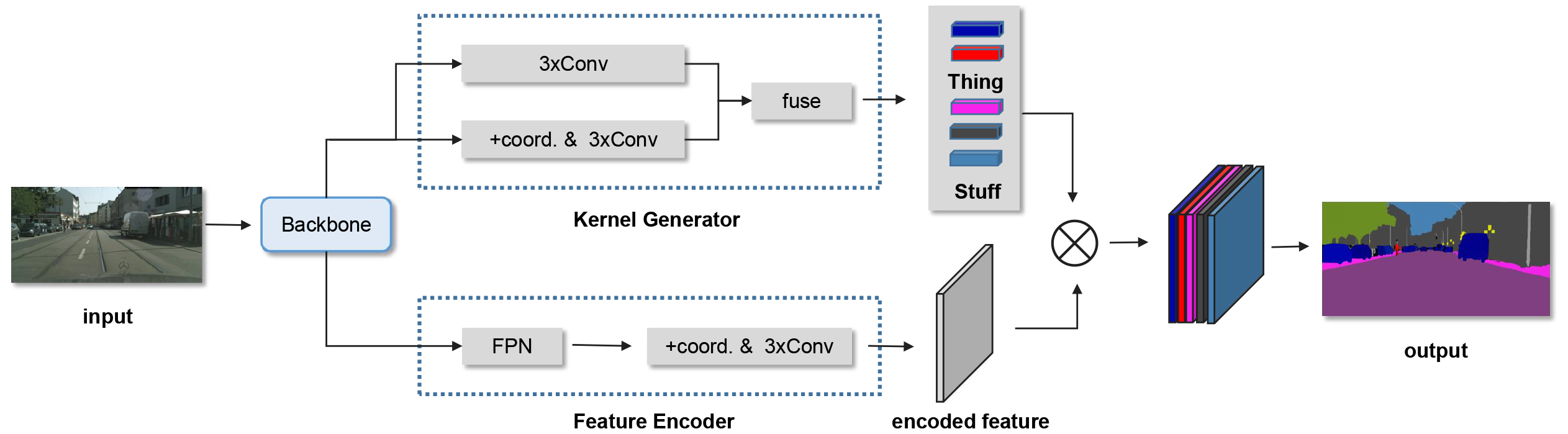}
\caption{ \textbf{Image Panoptic Segmentation Network.} We take PanopticFCN~\cite{li2021panopticfcn} as the image panoptic segmentation network to produce the segmentation predictions with a unified representation of object instances or stuff categories given an RGB image.}
\label{fig:method.panoptic segmentation}
\end{figure}

\subsection{Image Panoptic Segmentation Network}
\label{sec:method.panoptic segmentation}
Unlike VPSNet~\cite{kim2020vps}, which uses UPSNet~\cite{xiong2019upsnet} network based on Mask-RCNN~\cite{He_2017_ICCV}, we choose PanopticFCN~\cite{li2021panopticfcn} to obtain the image panoptic segmentation for every video frame. PanopticFCN~\cite{li2021panopticfcn} is a simple and efficient network for panoptic segmentation, which represents background stuff and foreground things in a unified fully convolutional pipeline.
As shown in Fig.~\ref{fig:method.panoptic segmentation}, PanopticFCN leverages the kernel generator to embed each stuff category and object instance into a specific kernel weight and generates the segmentation predictions. Compared with Mask-RCNN~\cite{He_2017_ICCV} used in VPS~\cite{kim2020vps}, PanopticFCN is a box-free method which is a memory and parameter-friendly network, and this makes our HybridTracker as lightweight as possible. We use the dice segmentation loss~\cite{milletari2016v} and position loss as the basic loss.

The dice segmentation loss is defined as follows:
\begin{equation} ~\label{equ:dice_raw}
{\mathcal L}_{\mathrm{seg}}=\sum_j\mathrm{Dice}(\mathbf{P}_j,\mathbf{Y}_j^{\mathrm{seg}})/(M+N), \\
\end{equation}
where $\mathbf{Y}_j^{\mathrm{seg}}$ is the $j$-th ground truth label of the prediction $\mathbf{P}_j$, while $\mathbf{P}_j=K_j\otimes \mathbf{F}^{\mathrm{e}}$. The $K_j$ is the kernel weights of $M$ things $K^{\mathrm{th}}$ and $N$ stuff $K^{\mathrm{st}}$, while $\mathbf{F}^{\mathrm{e}}\in\mathbb{R}^{C_\mathrm{e}\times W/4\times H/4}$ is the encoded feature. 

The position loss ${\mathcal L}_{\mathrm{pos}}$ is used to optimized the object centers with ${\mathcal L}_{\mathrm{pos}}^{\mathrm{th}}$ and stuff regions with ${\mathcal L}_{\mathrm{pos}}^{\mathrm{st}}$:

\begin{equation} ~\label{equ:loc}
\begin{aligned}
{\mathcal L}_{\mathrm{pos}}^{\mathrm{th}}=&\sum_i\mathrm{FL}(\mathbf{L}_i^{\mathrm{th}},\mathbf{Y}_i^{\mathrm{th}})/N_{\mathrm{th}}, \\
{\mathcal L}_{\mathrm{pos}}^{\mathrm{st}}=&\sum_i\mathrm{FL}(\mathbf{L}_i^{\mathrm{st}},\mathbf{Y}_i^{\mathrm{st}})/{W_i H_i}, \\
{\mathcal L}_{\mathrm{pos}} =& {\mathcal L}_{\mathrm{pos}}^{\mathrm{th}} + {\mathcal L}_{\mathrm{pos}}^{\mathrm{st}},
\end{aligned}
\end{equation}
where $\mathrm{FL}(\cdot,\cdot)$ denotes the Focal Loss~\cite{lin2018focal}. $\mathbf{L}_i^{\mathrm{th}}\in \mathbb{R}^{N_{\mathrm{th}}\times W_i\times H_i}$ is the object map, $\mathbf{L}_i^{\mathrm{st}}\in \mathbb{R}^{N_{\mathrm{st}} \times W_i\times H_i}$ is stuff map and ~$\mathbf{Y}_i^{\mathrm{st}}\in\left[0,1\right]^{N_{\mathrm{st}}\times W_i\times H_i}$ is the ground truth. The final basic loss is:
\begin{equation}
{\mathcal L_\mathrm{base}}=\lambda_{\mathrm{pos}}{\mathcal L}_{\mathrm{pos}} + \lambda_{\mathrm{seg}}{\mathcal L}_{\mathrm{seg}},
\end{equation}
where $\lambda_{\mathrm{pos}}$ is 1.0 and $\lambda_{\mathrm{seg}}$ is 4.0.

\begin{figure}
\centering
\includegraphics[width=\linewidth]{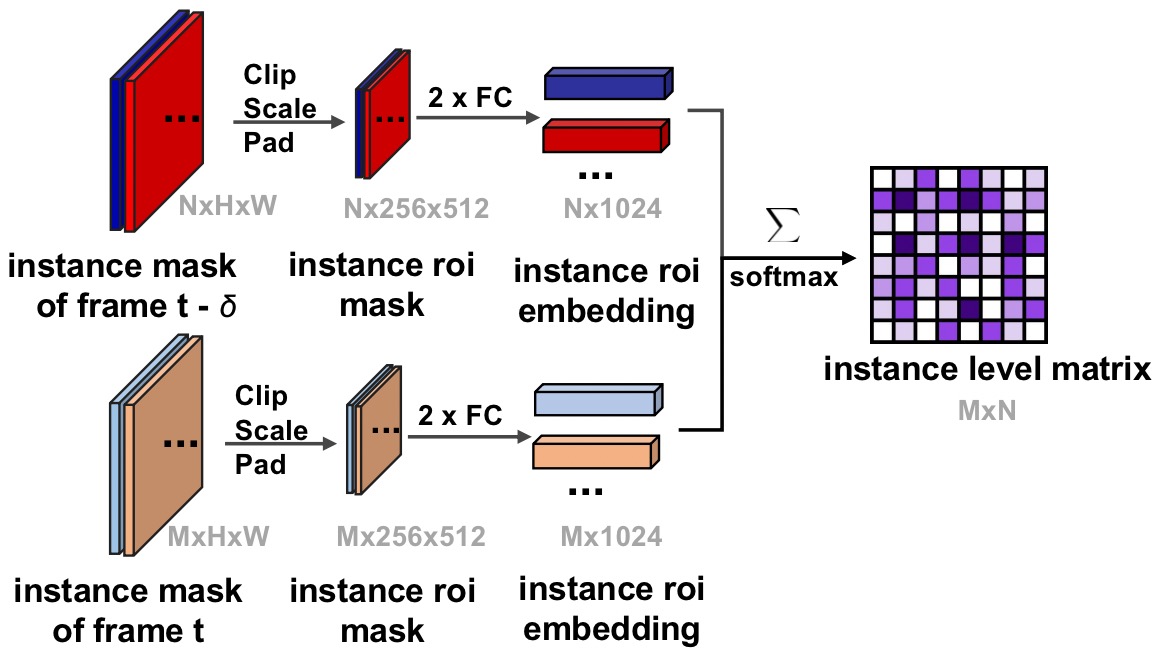}
\caption{\textbf{Instance Tracker Module.} Instance tracker takes each instance segmentation mask as input, crops, scales, and pads the predicted masks to the uniform size of masks, and two fully connected layers follow. We use the operation of $M_{mn}=\sum a_{mk} b_{mk}$ and $softmax$ to compute the instance similarity matrix.}
\label{fig:method.instance tracker}
\end{figure}

\subsection{Instance Tracker Module}
\label{sec:method.instance tracker}

After the images are fed into the image panoptic segmentation network, we can obtain each foreground object instance and background stuff, which can be used to learn the embedding of each instance for tracking between video frames. Unlike MaskTrack~\cite{yang2019video} which obtains a 28x28 mask based on the anchor, it is possible to directly use several fully connected layers to obtain the embedding of the corresponding instances. While PanopticFCN~\cite{li2021panopticfcn} obtains the mask of the original image, using the previous method directly may consume too much GPU memory and struggle to obtain good embedding. For an arbitrary two images frame $t-\delta$ and frame $t$, assume that frame $t-\delta$ has $m$ instances, we transform the binary mask of each instance into a bounding box, and crop the region enclosed by the bounding box into RoI feats, and then the RoI feature map is scaled to the shape of $256*512$, the empty regions are padded by 0, which ensures no deformation occurred when the RoI mask obtained. We get $m*1024$ embedding by two fully connected layers and yield an embedding of $n*1024$ for frame $t$, which has $n$ instances. To calculate the similarity of each two instances of the two images, we multiply the embedding of each instance two by two to obtain $m*n$ values, which forms the instance matching matrix $M_{mn}=\sum a_{mk} b_{mk}$. 
The instance tracker is detailed in Fig.~\ref{fig:method.instance tracker}. We can get the matching indexes simply through the operation of $torch.argmax$. We also used bipartite graph matching~\cite{yu2020learning} to solve the instance correspondence and found less improvement, probably because the learned embedding is robust enough that it may not cause ambiguous matches.


\noindent \textbf{Instance-Wise Differentiable Matching layer based on Instance Embedding.}
\label{sec:method.instance tracker.Differentiable Matching layer}
In the instance tracker, the instance embedding is trained to track the elements between frames. Existing methods use triplet loss~\cite{Ye2021SuperPlane} or contrast loss~\cite{woo2021learning} to pull the same instance as close as possible in feature space and push different instances as far as possible. These methods identify nearest neighbor matches in feature space to establish instance correspondences, which are non-differentiable and unstable during training.

\begin{figure}
\centering
\includegraphics[width=\linewidth]{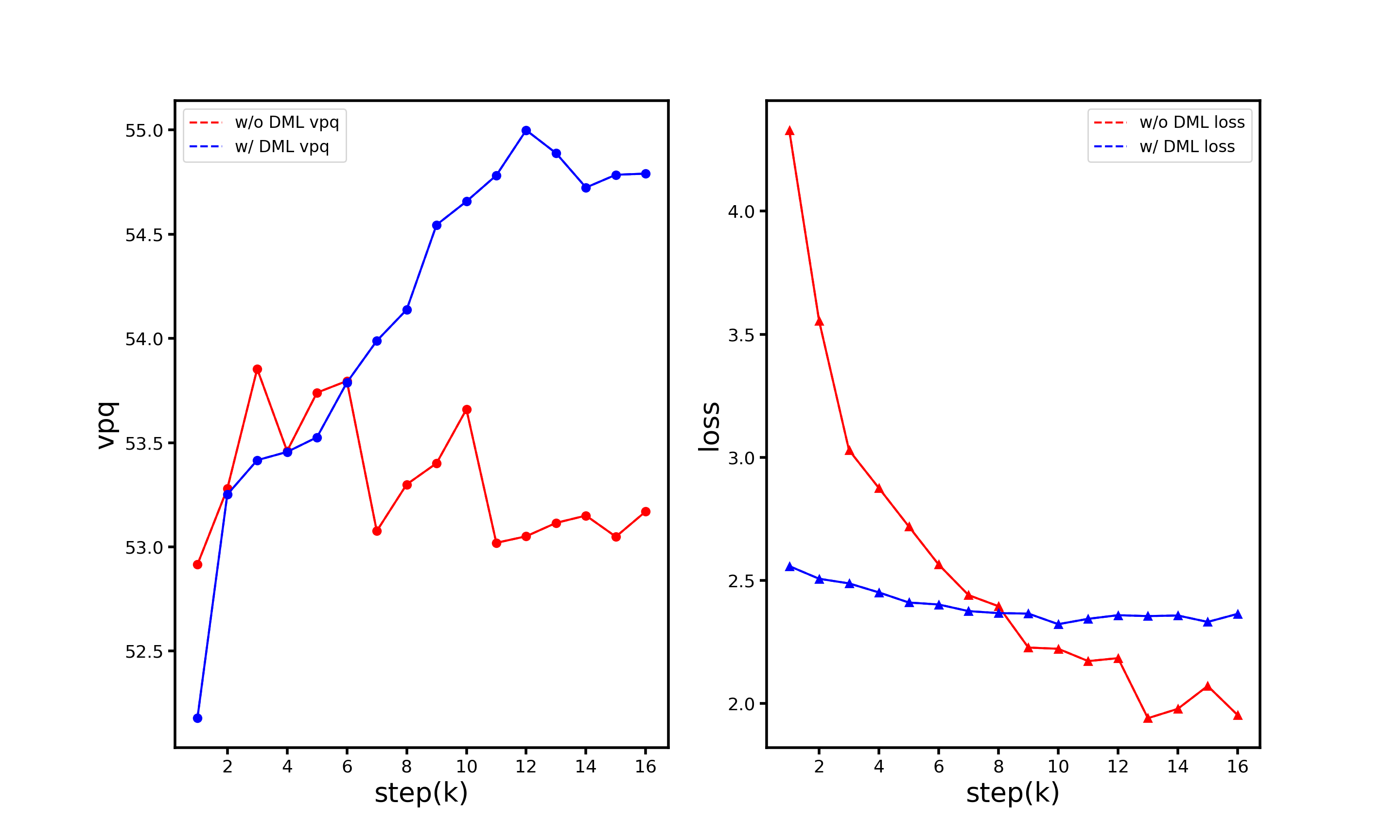}
\caption{\textbf{Training and testing of instance tracker w/ VS w/o the differentiable matching layer (DML).} DML helps instance tracker converge faster and more stable with higher VPQs.}
\label{fig:method.Training of Instance Tracker}
\end{figure}

To address this issue, inspired by CAPS~\cite{wang2020learning}, we propose to use an instance-wise differentiable matching layer (DML) for the training of instance embedding. Given a set of frame $t-\delta$ and frame $t$, the panoptic segmentation network with shared weights mentioned above obtains the embedding vectors $M$ and $N$ respectively. To compute the correspondence of the instance $i$ in frame $t-\delta$ to the frame $t$, we correlate the instance embedding, denoted $\mathbf{M}_\mathbf{i}$. Then, we use a 2D softmax operation to obtain the distribution of the different instances of frame $t$, which represents the probability that the instance of image frame $t$ corresponds to the instance $i$:  
\begin{equation}
    p(\mathbf{j}|\mathbf{i}, \mathbf{M}, \mathbf{N}) = \frac{\exp{(\mathbf{M}(\mathbf{i})^\text{T} \mathbf{N}(\mathbf{j}))}}{\sum_{\mathbf{x}\in \mathbf{t}}\exp{(\mathbf{M}(\mathbf{i})^\text{T}\mathbf{N}(\mathbf{x}))}},
    \label{eq:prob}
\end{equation}
where x is the instance indexes of frame $t$.
Then the predicted target $\hat{\mathbf{j}}$ can be computed as the expectation of the distribution:
\begin{equation}
     \hat{\mathbf{j}}= h_{t-\delta \rightarrow t}(\mathbf{i}) = \sum_{\mathbf{j} \in \mathbf{t}} \mathbf{j} \cdot p(\mathbf{j}|\mathbf{i}, \mathbf{M}, \mathbf{N}) .
\end{equation}
We use the cross-entropy loss to optimize the ground-truth target $j$ and the predicted target $\hat{\mathbf{j}}$, which is stable and easy to converge during the training process. The matching loss can be defined as follows:
\begin{equation}
    \mathcal L_{\mathrm{match}} = \frac{1}{M}\sum^M_{m = 1} -{(y\log(p) + (1 - y)\log(1 - p))},
\end{equation}
where $p$ represents the probability of correspondence between the ground-truth target $j$ and the predicted target $\hat{\mathbf{j}}$, i.e., $p(\mathbf{j}|\mathbf{i},\mathbf{M}, \mathbf{N})$. $y$ is binary indicator if predicted target $\hat{\mathbf{j}}$ is the correct classification for ground-truth target $j$.

\subsection{Pixel Tracker Module}
\label{sec:method.pixel tracker}

\begin{figure}
\centering
\includegraphics[width=\linewidth]{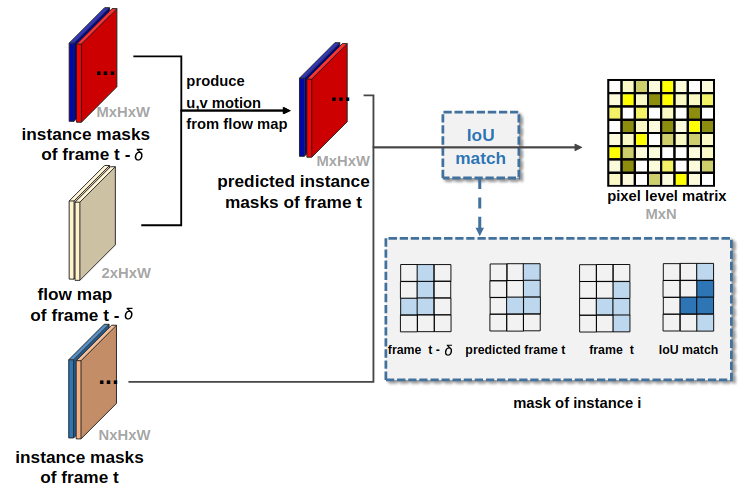}
\caption{ \textbf{Pixel Tracker Module.} We calculate the dice coefficient of different instances from the frame $t-\delta$ and frame $t$ given the optical flow map, to form the position-aware similarity matrix.}
\label{fig:method.pixel tracker}
\end{figure} 

Pixel tracker can be considered as a tracker that maintains consistency in terms of the spatial position of the instances. 
Unlike CenterTrack~\cite{zhou2020tracking} which exploits sparse objects' center to track multi-objects, we propose to track each instance by dense pixel motion. In pixel tracker (see Fig.~\ref{fig:method.pixel tracker}), the images of two adjacent frames are fed into the optical flow estimation network, such as RAFT~\cite{teed2020raft}, to obtain the inter-frame motion information. Take the instance $i$ in frame $t-\delta$ as an example, we multiply the binary mask of instance $i$ and flow map to get the motion of each pixel of instance $i$. Then, we add the coordinates of each pixel of the instance and the motion shift pixel by pixel to get the mask of each instance in the next frame. We can calculate the dice coefficient~\cite{milletari2016v} values between the $m$ instances in the frame $t-\delta$ and the $n$ instances in the frame $t$ as the position similarity. The position similarity can be defined as follows:
\begin{equation}
Sim_{pos}=\frac{2\sum_{i}^{N}p_{i}g_{i}}{\sum_{i}^{N}p_{i}^{2}+\sum_{i}^{N}g_{i}^{2}},
\end{equation}
where $p_i\in{P}$ means the predicted binary segmentation mask of instance $i$ and $g_i\in{G}$ means the predicted binary mask of instance $i$. Then we can form an $m*n$ IoU correlation matrix, where $m$ means the number of segmentation masks of the frame $t-\delta$, and $n$ means the number of masks of the frame $t$. 

\subsection{Fusion Module}
We take the two correlation matrices obtained by pixel tracker and instance tracker as input, feed them to a simple 1x1 Conv layer, and fuse them into a joint correlation matrix by adding them pixel by pixel to obtain the matching relationship between the two frames. The matching of instance IDs is obtained by taking the id of the largest column row by row, where matches probabilities less than a certain threshold, such as 1e-5, are also removed, and a 1-to-1 matching constraint is added. For example, row i matches column j --$>$ (i, j), and the corresponding column j also should match row i --$>$ (j, i), which is considered a correct match. Matches that do not satisfy this constraint will be removed.  Due to the large gap between the instance tracker and the pixel tracker, a simple fusion will only give intermediate results, while using our fusion module will generate better results, which also shows the effectiveness of our HybridTracker. 

\begin{figure}
\centering
\includegraphics[width=\linewidth]{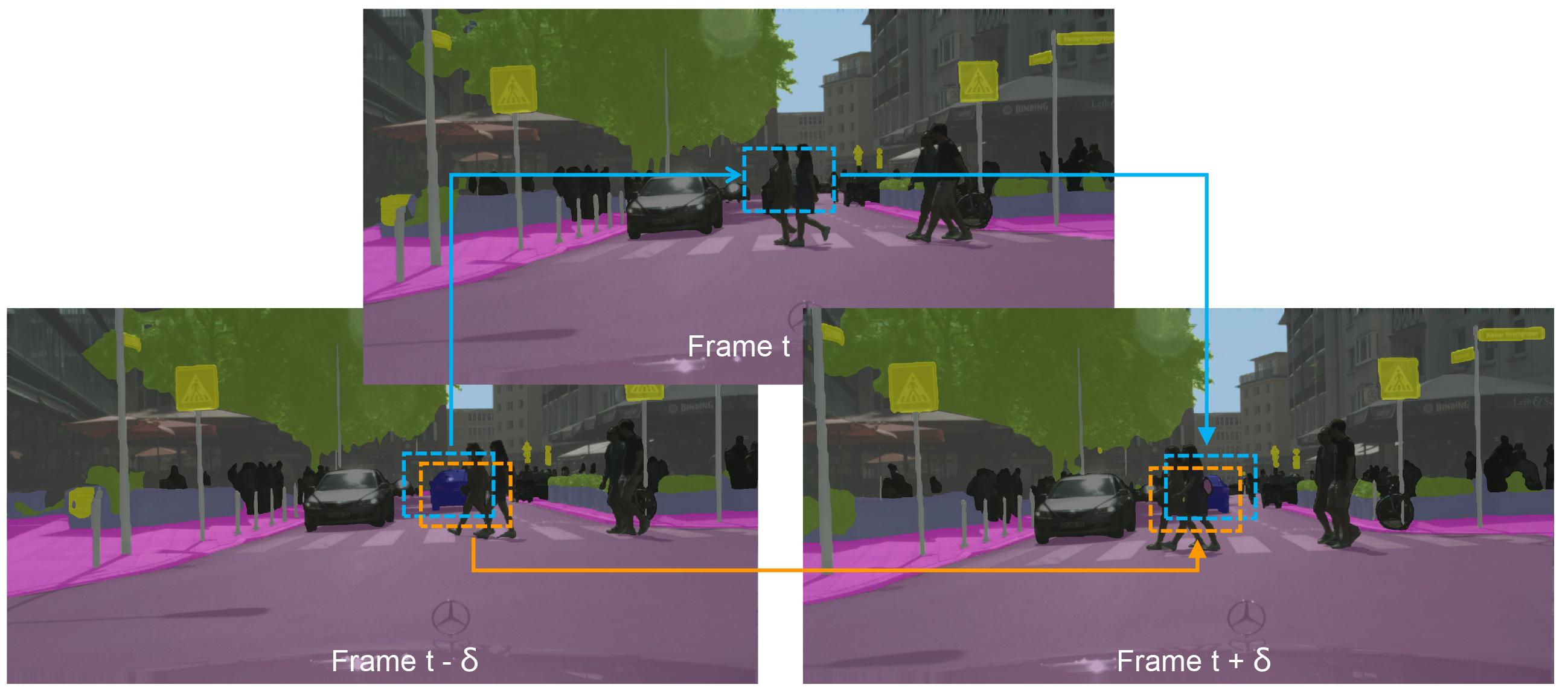}
\caption{ \textbf{Illustration of temporal consistency constraint}. The car in the blue line box is incorrectly segmented in frame $t$ due to occlusion, resulting in missing segmentation results. The blue line indicates that directly using the adjacent frame information may cause the tracking between frame $t-\delta$ and frame $t+\delta$ to fail. In comparison, the orange color line can address this limitation by the temporal consistency constraint.}
\label{fig:method.temporal consistency illustration}
\end{figure}

\subsection{Mutual Check and Temporal Consistency Constraints}
\begin{figure*}[!htbp]
  \centering
  \subfigure[Instance Tracker.]{
  \begin{minipage}[]{0.23\linewidth}
        \centering
        \includegraphics[width=4.40cm]{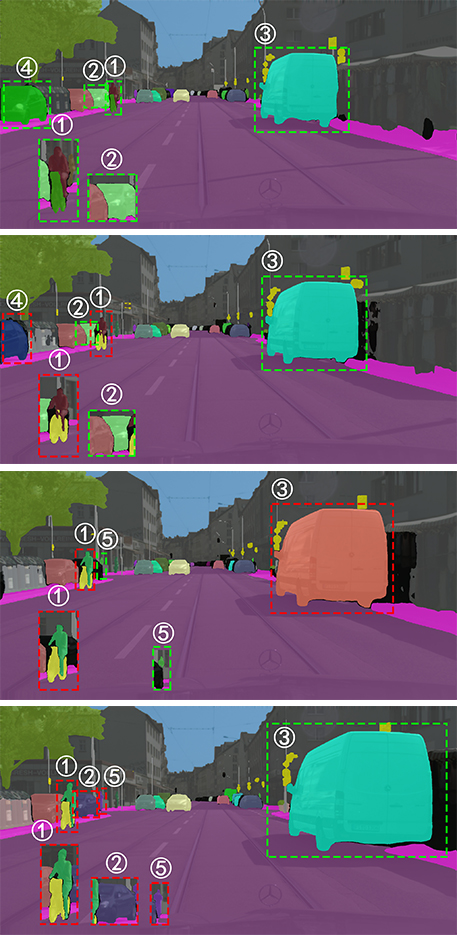}\\
    \end{minipage}%
    \label{fig:comparison Instance Tracker}
  }
  \subfigure[Pixel Tracker.]{
  \begin{minipage}[]{0.23\linewidth}
        \centering
        \includegraphics[width=4.40cm]{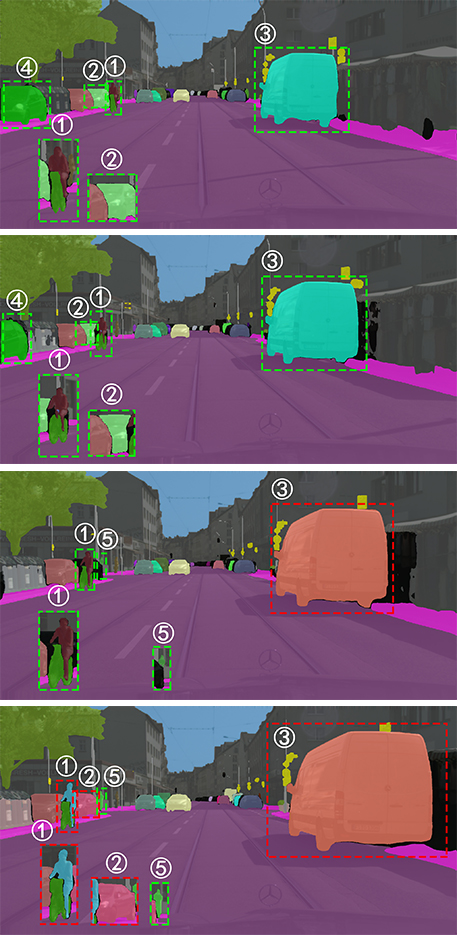}\\
    \end{minipage}%
    \label{fig:comparison val Pixel Tracker}
  }
  \subfigure[HybridTracker.]{
  \begin{minipage}[]{0.23\linewidth}
        \centering
        \includegraphics[width=4.40cm]{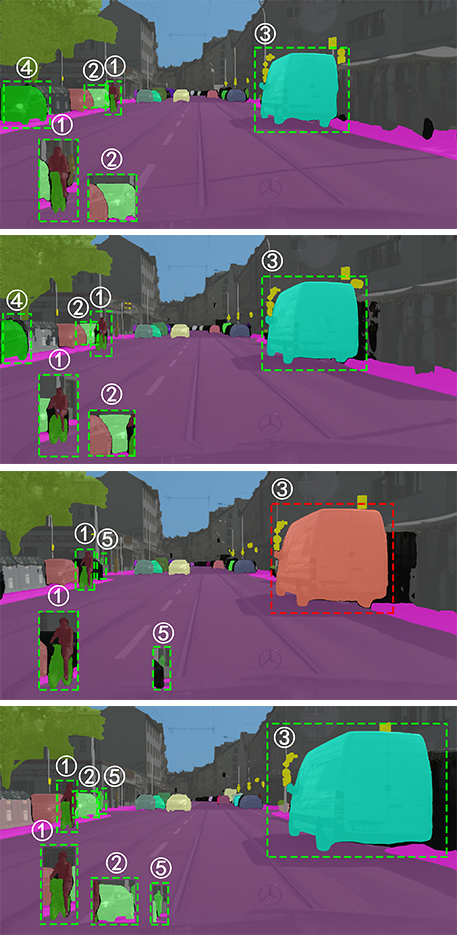}\\
    \end{minipage}%
    \label{fig:comparison Hybrid Tracker}
  }
  \subfigure[HybridTracker + w/ Temporal.]{
  \begin{minipage}[]{0.23\linewidth}
        \centering
        \includegraphics[width=4.40cm]{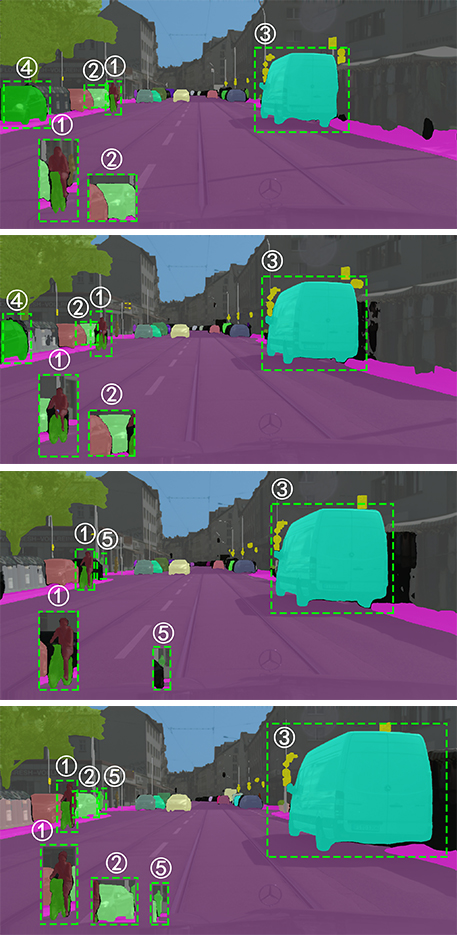}\\
    \end{minipage}%
    \label{fig:comparison Hybrid Tracker w/ Temporal}
  }
  \caption{\textbf{Illustration of our different trackers on Cityscape-VPS dataset.} Each column with the same marker indicates a tracking instance. The green box indicates correct tracking,
while the red box indicates false tracking. Instance Tracker can handle large objects with minor contour deformation but struggles to handle small objects. In contrast, Pixel Tracker can handle small objects with limited motion, rather than irregular motion (\whiteding{4} and \whiteding{5}). HybridTracker is expert in settling small objects with large motion (\whiteding{1} and \whiteding{2}). The temporal consistency constraints help track small and large objects with occlusion and inter-frame jump due to deformation (\whiteding{3}).}
\label{fig:comparison ablation study hybridtracker temporal constraints}
\vspace{-1em}
\end{figure*}

Although our HybridTracker can effectively solve the unstable tracking phenomenon of the single tracker, shown in Fig.~\ref{fig:comparison Hybrid Tracker}, we found that we still cannot handle the complex deformation of some objects well. For this reason, we propose the mutual check which aims to remove inconsistent matches between two instances, and the temporal consistency constraints~\cite{CVPR2019_CycleTime} which constrain the consistency of tracking between multiple frames in inference. 
Assume that $\mathcal{T}$ is a differentiable operation $x^p_t \mapsto {x}^p_s$, which measures the similarity of $x^p_s$ and $x^p_t$, where $s$ and $t$ indicate timestamp, $p$ can be regarded as the spatial location of pixels or the feature of the instances. We can apply $\mathcal{T}$ from $t-\delta$ to $t$ and from $t$ to $t+\delta$, and the temporal consistency can be established: 
\begin{equation}
\mathcal{T}(x^p_{t-\delta}, x^p_{t+\delta}) = \mathcal{T}(x^p_{t-\delta},x^p_{t}) \mathcal{T}(x^p_{t}, x^p_{t+\delta}).
\end{equation}

An obvious example illustrates that the temporal consistency constraint can robustly settle the issues of multi-object occlusion and missing segmentation of individual instances in the video sequence, shown in Fig.~\ref{fig:method.temporal consistency illustration}. 
In inference, since the car in frame $t+\delta$ fails to find a matching instance in frame $t$, we obtain the frame $t-\delta$ stored in memory and use our tracker (instance tracker or pixel tracker or hybrid tracker) to obtain the similarity matrix between frame $t-\delta$ and $t+\delta$. If the similarity value is greater than the specific threshold, we will adopt temporal matches. Otherwise, the new instance id will be set. By the way, the temporal consistency constraint should be further applied backward from time $t+\delta$ to $t$ and from $t$ to $t-\delta$. This temporal consistency constraint can also be used for other sequence problems to ensure consistency.

\section{Implementation Details}
\label{sec:appendix.Implementation Details}
We first pre-train the image panoptic segmentation network, PanopticFCN~\cite{li2021panopticfcn} without tracking functionality on the Cityscapes-VPS and VIPER datasets
and implement our models with the Pytorch-based toolbox Detectron2~\footnote{https://github.com/facebookresearch/detectron2}. Following VPSNet-Track~\cite{kim2020vps}, we adopt the same multi-scale inference scaling equal to {0.5, 0.625, 0.6875, 0.8125, 0.875, 1}. The resolution of the Cityscapes image is $1024\times 2048$, while the VIPER image is $1080\times 1920$. We optimize the network with an initial rate of 0.0025 on 4 GeForce GTX TITAN X GPUs, where each mini-batch has four images. We use the SGD optimizer with weight decay 1e-4 and momentum 0.9. Then we finetune the instance tracker with the differentiable matching layer on the Cityscapes and VIPER datasets. For finetuning our instance tracker, we optimize the network with initial rate 1e-4 and 2 GeForce GTX TITAN X GPUs, where each mini-batch has 16 images. The pixel-tracker does not need training, which loads the pre-trained model weight from the optical flow network. We compose the instance tracker and the pixel tracker to the hybrid tracker to fuse the matching matrix.  We also add a mutual check and temporal consistency constraints during inference.

\begin{table*}[]
\centering
\caption{\textbf{Quantitative comparisons of video panoptic segmentation on Cityscapes-VPS validation (top) and test (bottom) Set. Our methods generally outperforms VPSNet-FuseTrack~\cite{kim2020vps} and SiamTrack~\cite{woo2021learning}.} Each cell contains VPQ / VPQ\textsuperscript{Th} / VPQ\textsuperscript{St} scores. The best results are highlighted in boldface. \textbf{Note} that the configuration of our method is consistent with VPSNet-Track. }
\resizebox{\textwidth}{!}{%
\begin{adjustbox}{max width=\textwidth}
\begin{tabular}{l|c|c|c|c| c|c}
\hline
{Methods} &\multicolumn{4}{c|}{Temporal window size} 
                & \multirow{2}{*}{VPQ} & \multirow{2}{*}{FPS} \\
\cline{2-5} {on \textbf{Cityscapes-VPS \textit{val}}} & k = 0 & k = 5 & k = 10 & k = 15 &  \\
\hline
VPSNet-Track  & 
			63.1 / 56.4 / 68.0 & 
            56.1 / 44.1 / 64.9 & 
            53.1 / 39.0 / 63.4 &
            51.3 / 35.4 / 62.9 &   
            55.9 / 43.7 / 64.8 &
            4.5\\

VPSNet-FuseTrack \quad \quad \quad   & 
            64.5 / 58.1 / 69.1 & 
            57.4 / 45.2 / 66.4 &
            54.1 / 39.5 / 64.7 &   
            52.2 / 36.0 / 64.0 & 
            57.2 / 44.7 / 66.6 &
            1.3\\

SiamTrack &
64.6 / 58.3 / 69.1 &
57.6 / 45.6 / 66.6 &
54.2 / 39.2 / 65.2 &
52.7 / 36.7 / 64.6 &
57.3 / 44.7 / 66.4 &
4.5\\
\hline
Instance Tracker  & 
			65.6 / 60.0 / 69.7 & 
            55.1 / 39.3 / 66.6 & 
            51.4 / 32.5 / 65.1 &
            49.3 / 28.6 / 64.3 & 
            55.3 / 40.1 / 66.4 &
            5.1 \\



            
Pixel Tracker  & 
			65.6 / 60.0 / 69.7 & 
            58.8 / 48.0 / 66.6 & 
            55.4 / 42.1 / 65.1 &
            53.2 / 38.0 / 64.3 & 
            58.2 / 47.0 / 66.4 &
            3.8 \\
            
            
HybridTracker (Instance + Pixel) & 
			65.6 / 60.0 / 69.7 & 
            58.9 / 48.3 / 66.6 & 
            55.6 / 42.5 / 65.1 &
            53.4 / 38.5 / 64.3 &  
            58.4 / 47.3 / 66.4 &
            3.6 \\
            
HybridTracker + w/ Temporal \quad \quad \quad   & 
            \textbf{65.6} / 60.0 / 69.7 & 
            \textbf{59.0} / 48.5 / 66.6 &
            \textbf{55.7} / 42.8 / 65.1 &   
            \textbf{53.6} / 38.9 / 64.3 & 
            \textbf{58.5} / 47.5 / 66.4&
            3.3\\

\hline
\end{tabular}
\end{adjustbox}
}
\\
\centering
\resizebox{\textwidth}{!}{%
\begin{adjustbox}{max width=\textwidth}
\begin{tabular}{l|c|c|c|c| c|c}
\hline
{Methods} &\multicolumn{4}{c|}{Temporal window size} 
                & \multirow{2}{*}{VPQ} & \multirow{2}{*}{FPS} \\
\cline{2-5} {on \textbf{Cityscapes-VPS \textit{test}}} & k = 0 & k = 5 & k = 10 & k = 15 &  \\
\hline
VPSNet-Track  & 
			63.1 / 58.0 / 66.4 & 
            56.8 / 45.7 / 63.9 & 
            53.6 / 40.3 / 62.0 &
            51.5 / 35.9 / 61.5 &     
            56.3 / 45.0 / 63.4 &
            4.5 \\

VPSNet-FuseTrack \quad \quad \quad   & 
			64.2 / 59.0 / 67.7 & 
            57.9 / 46.5 / 65.1 & 
            54.8 / 41.1 / 63.4 &
            52.6 / 36.5 / 62.9 &   
            57.4 / 45.8 / 64.8 &
            1.3 \\

SiamTrack &
            63.8 / 59.4 / 66.6 &
            58.2 / 47.2 / 65.9 &
            56.0 / 43.2 / 64.4 &
            \textbf{54.7} / 40.2 / 63.2 &
            57.8 / 47.5 / 65.0 &
            4.5 \\
\hline
Instance Tracker & 
			61.7 / 56.2 / 65.2 & 
            55.0 / 43.0 / 62.6 & 
            52.2 / 38.1 / 61.2 &
            50.3 / 34.4 / 60.5 &
            54.8 / 42.9 / 62.4 &
            5.1 \\

Pixel Tracker & 
			64.2 / 58.9 / 67.7 & 
            58.4 / 48.0 / 65.1 & 
            55.6 / 43.3 / 63.4 &
            53.8 / 39.7 / 62.9 &     
            58.0 / 47.5 / 64.8 &
            3.8 \\

HybridTracker (Instance + Pixel) & 
			64.3 / 58.9 / 67.7 & 
            58.6 / 48.5 / 65.1 & 
            55.8 / 43.9 / 63.4 &
            54.1 / 40.2 / 62.9 &
            58.2 / 47.9 / 64.8 &
            3.6 \\

HybridTracker + w/ Temporal & 
			\textbf{64.3} / 59.0 / 67.7 & 
            \textbf{59.0} / 49.2 / 65.1 & 
            \textbf{56.3} / 45.0 / 63.4 &
            54.5 / 41.4 / 62.9 &  
            \textbf{58.5} / 48.7 / 64.8 &
            3.3\\
            

\hline
\end{tabular}
\end{adjustbox}
}
    
\label{tab:cityvps_vpq val and test}
\end{table*}       
 
\begin{table*}[]
\centering
\caption{\textbf{Quantitative comparisons of video panoptic segmentation on VIPER. Our methods generally outperforms VPSNet-FuseTrack~\cite{kim2020vps} and SiamTrack~\cite{woo2021learning}.} Each cell contains VPQ / VPQ\textsuperscript{Th} / VPQ\textsuperscript{St} scores. The best results are highlighted in boldface. \textbf{Note} that the configuration of our method is consistent with VPSNet-Track. }
\resizebox{\textwidth}{!}{%

\begin{adjustbox}{max width=\textwidth}
\begin{tabular}{l|c|c|c|c| c|c}
\hline
{Methods} &\multicolumn{4}{c|}{Temporal window size} 
                & \multirow{2}{*}{VPQ} & \multirow{2}{*}{FPS} \\
 \cline{2-5} {on \textbf{VIPER}} & k = 0 & k = 5 & k = 10 & k = 15 &  \\
\hline
VPSNet-Track  & 
		    48.1 / 38.0 / 57.1 & 
            49.3 / 45.6 / 53.7 & 
            45.9 / 37.9 / 52.7 &
            43.2 / 33.6 / 51.6 &  
            46.6 / 38.8 / 53.8 &
            5.1 \\
            
VPSNet-FuseTrack      & 
			49.8 / 40.3 / 57.7 & 
            51.6 / 49.0 / 53.8 & 
            47.2 / 40.4 / 52.8 &
            45.1 / 36.5 / 52.3 &   
            48.4 / 41.6 / 53.2 &
            1.6 \\
            SiamTrack &
            51.1 / 42.3 / 58.5 &
            \textbf{53.4} / 51.9 / 54.6 &
            49.2 / 44.1 / 53.5 &
            47.2 / 40.3 / 52.9 &
            50.2 / 44.7 / 55.0 &
            5.1 \\
\hline
Instance Tracker  & 
			55.1 / 51.2 / 58.0 & 
            45.7 / 30.4 / 57.5 & 
            43.8 / 26.2 / 57.3 &
            41.5 / 21.5 / 57.0 &  
            46.5 / 32.3 / 57.5& 
            5.7 \\
            
Pixel Tracker  & 
			\textbf{55.1} / 51.2 / 58.0 & 
            52.2 / 45.4 / 57.5 & 
            50.9 / 42.4 / 57.3 &
            49.5 / 39.9 / 57.0 &
            51.9 / 44.7 / 57.5 &
            3.8 \\


HybridTracker (Instance + Pixel) & 
            55.0 / 51.2 / 58.0 & 
            52.4 / 45.7 / 57.5 & 
            51.0 / 42.8 / 57.3 &
            49.8 / 40.4 / 57.0 &  
            52.1 / 45.1 / 57.5 &
            3.6 \\

HybridTracker + w/ Temporal & 
			55.0 / 51.2 / 58.0 & 
            52.6 / 46.2 / 57.5 & 
            \textbf{51.5} / 43.9 / 57.3 &
            \textbf{50.4} / 42.0 / 57.0 & 
            \textbf{52.4} / 45.8 / 57.5 &
            3.5 \\


            
\hline
\end{tabular}
\end{adjustbox}
}

\label{tab:viper_vpq}
\end{table*}

\begin{figure*}[!htbp]
  \centering
  \subfigure[VPSNet-FuseTrack (val).]{
  \begin{minipage}[]{0.23\linewidth}
        \centering
        \includegraphics[width=4.45cm]{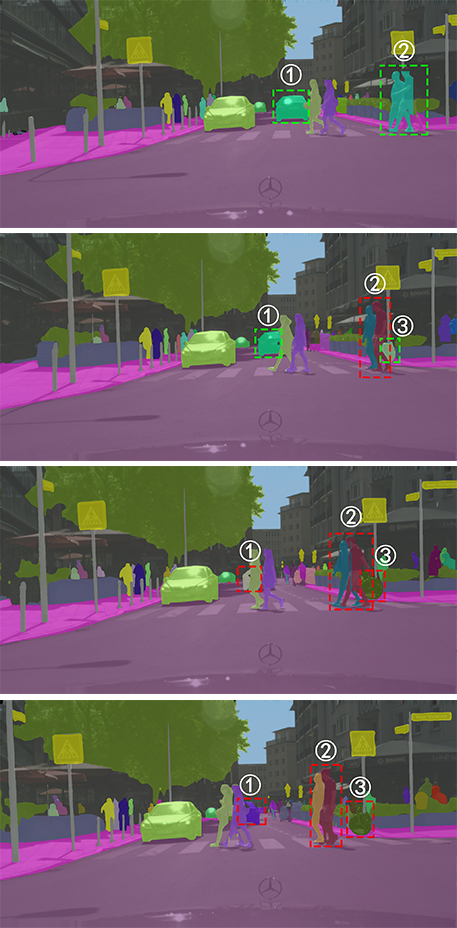}\\
    \end{minipage}%
    \label{fig:comparison val vps}
  }
  \subfigure[HybridTracker (val).]{
  \begin{minipage}[]{0.23\linewidth}
        \centering
        \includegraphics[width=4.45cm]{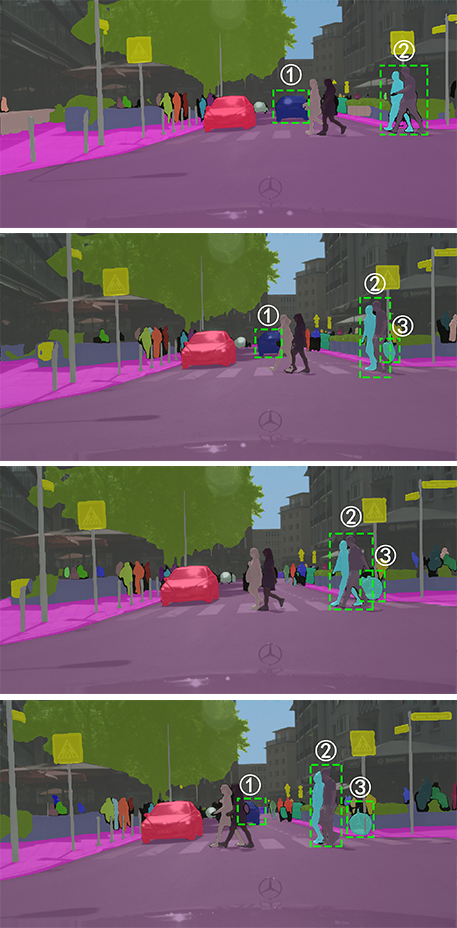}\\
    \end{minipage}%
    \label{fig:comparison val hybrid(ours)}
  }
  \subfigure[VPSNet-FuseTrack (test).]{
  \begin{minipage}[]{0.23\linewidth}
        \centering
        \includegraphics[width=4.45cm]{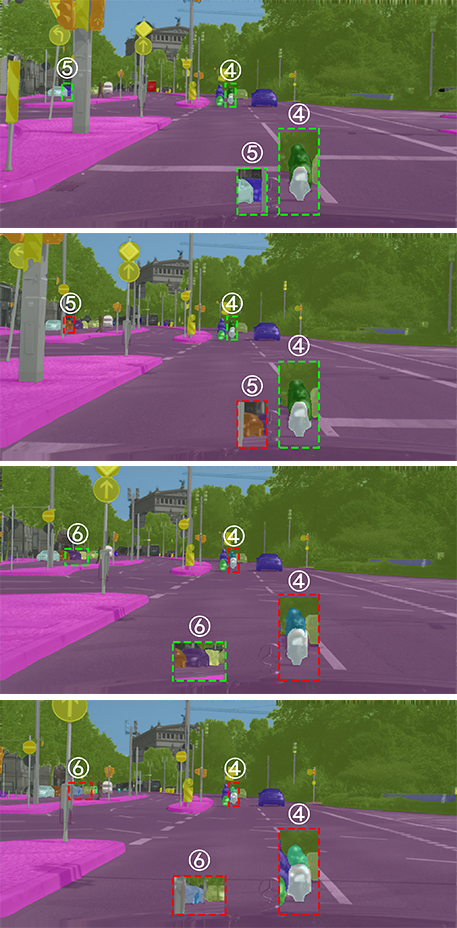}\\
    \end{minipage}%
    \label{fig:comparison test vps}
  }
  \subfigure[HybridTracker (test).]{
  \begin{minipage}[]{0.23\linewidth}
        \centering
        \includegraphics[width=4.45cm]{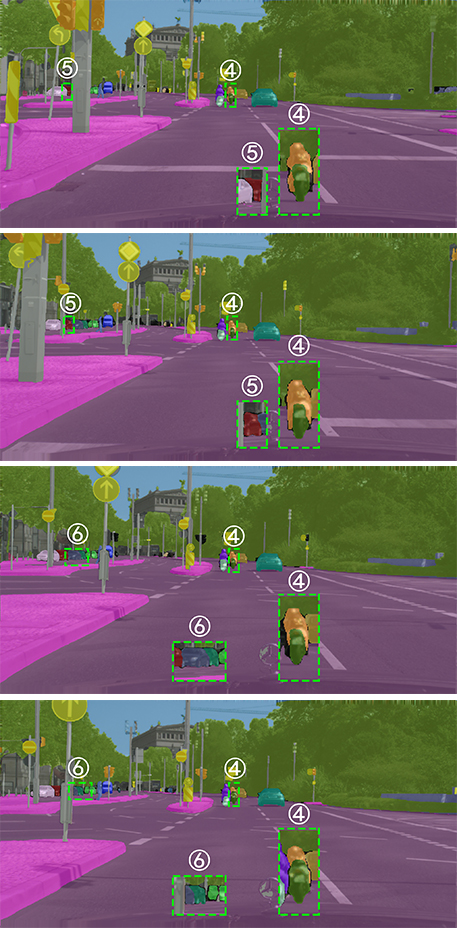}\\
    \end{minipage}%
    \label{fig:comparison test hybrid(ours)}
  }
  \caption{\textbf{Comparison results of our methods with VPSNet-FuseTrack~\cite{kim2020vps} on Cityscape-VPS val and test dataset}. Each column with the same marker indicates a
tracking instance. The green box indicates correct tracking,
while the red box indicates false tracking. Compared with VPSNet-FuseTrack, HybridTrack can robustly handle multiple object occlusions ( \whiteding{1} and \whiteding{3}) and similar appearance but different identities (\whiteding{2}) on the val dataset and maintain robust tracking on the test dataset even for small objects such as a person on a motorcycle (\whiteding{4}), a car (\whiteding{6}), and a car obscured by a utility pole (\whiteding{5}).}
  \label{fig:qualtitative result compared with vpsnet}
\end{figure*}


\section{Experiment}
\label{sec:exp}

\subsection{Datasets}
We extensively perform experiments on two VPS datasets, including Cityscapes-VPS datasets and VIPER datasets. For fail comparison, we follow VPS~\cite{kim2020vps} and use its public train/val/test split of these datasets. Each video of the Cityscapes-VPS consists of thirty consecutive frames with every five frames having ground truth annotation, a total of 6 frames are annotated. VIPER is another high-quantity and quality video panoptic segmentation benchmark. Following VPS~\cite{kim2020vps}, We select the first 60 frames of 10 videos of the day scenarios as the validation set, with a total of 600 images.



\subsection{Experimental Details}
We consider Cityscapes-VPS as a primary evaluation dataset as the dataset is relatively small, which allows us to quickly validate our various experimental settings, and then we analogize our learned knowledge to the VIPER dataset, which also shows consistent results.  

We selected three instance-based video panoptic segmentation methods as comparison, including VPSNet-Track, VPSNet-FuseTrack~\cite{kim2020vps} and SiamTrack~\cite{woo2021learning}. VPSNet-Track is a video panoptic segmentation baseline that adds MaskTrack head~\cite{yang2019video} on top of the single image panoptic segmentation model, UPSNet~\cite{xiong2019upsnet}. VPSNet-FuseTrack based on VPSNet-Track, adds temporal feature aggregation and fusion module to improve the performance of video panoramic segmentation. SiamTrack~\cite{woo2021learning} finetunes VPSNet-Track with the pixel-tube matching loss and the contrast loss and is only slightly better than VPSNet-FuseTrack. We mainly compare with VPSNet-FuseTrack~\footnote{https://github.com/mcahny/vps} for SiamTrack's source code is not publicly available. The VPQ results are from SiamTrack~\cite{woo2021learning}.


To better evaluate the joint tracking network, 
we experiment with the proposed instance tracker, pixel tracker, and hybrid tracker respectively. We additionally add a mutual check and temporal consistency constraints to the single and hybrid tracker. 
The VPQ metric~\cite{kim2020vps} is adopted to evaluate the spatial-temporal consistency between the ground truth and predicted video panoptic segmentation. 




\subsection{Main Results}
\subsubsection{\textbf{Cityscapes-VPS}}
Tab.~\ref{tab:cityvps_vpq val and test} summarizes video panoptic segmentation (VPQ) results of all methods on the Cityscapes-VPS val and test dataset. We observed that our Instance Tracker is on par with Track, while Pixel Tracker is better than VPSNet-FuseTrack, which makes our full HybridTracker outperform VPSNet-FuseTrack, and achieve faster inference (\textbf{3.6FPS} vs. 1.3FPS) and fewer parameters (\textbf{170.8MB} vs. 214.9MB).  
Furthermore, HybridTracker with temporal consistency constraint has further improvements, outperforming VPSNet-FuseTrack with \textbf{+1.3\% VPQ} gain for the val dataset and \textbf{+1.1\% VPQ} gain for the test dataset.
\textbf{Note} that all our model does not fuse feature from consecutive frames like VPSNet-FuseTrack~\cite{kim2020vps}.  In addition, we combine better panoptic segmentation results from VPSNet-FuseTrack with our pixel tracker on the Cityscpaes-VPS val dataset to achieve better VPQ scores, compared with VPSNet-FuseTrack.
Fig.~\ref{fig:qualtitative result compared with vpsnet} demonstrates that compared with VPSNet-Fuse
Track, HybridTrack can robustly handle multiple object occlusions ( \whiteding{1} and \whiteding{3}) and similar appearance but different identities (\whiteding{2}) 
and maintain robust tracking 
even for small objects such as a person on a motorcycle (\whiteding{4}), a car (\whiteding{6}), and a car obscured by a utility pole (\whiteding{5}). The reason why HybridTracker does not have a big improvement over Pixel Tracker is that Pixel Tracker is relatively robust, and the proportion of bad cases is not too large. In addition, there is still a big gap between Instance Tracker and Pixel Tracker in performance, and simple fusion can only get intermediate results, which also reflects the effectiveness of our HybridTracker.

\subsubsection{\textbf{VIPER}}
Tab.~\ref{tab:viper_vpq} summarizes the VPQ results of all the methods on VIPER datasets, and we observe a consistent tendency. Compared with VPSNet-FuseTrack, our HybridTracker model with pixel and instance tracker module achieves much higher scores (\textbf{52.1VPQ} vs. 48.4 VPQ) using the learned RoI feature matching algorithm and the spatial position correlation cues. Our HybridTracker with the mutual check and temporal consistency constraints achieves the best performance by improving \textbf{+4.0\%VPQ} over VPSNet-FuseTrack, and runs at a much faster inference speed (2x).

\subsection{Ablation Study}
\label{subsec: experiments: ablation study}
\subsubsection{\textbf{Analysis on HybridTracker}}
We evaluate the VPQ performance of different variants of our methods. From Tab.~\ref{tab:cityvps_vpq val and test} and Tab.~\ref{tab:viper_vpq}, the proposed instance tracker is slightly worse than VPSNet-FuseTrack~\cite{kim2020vps}. To be fair to the existing method, VPSNet, our instance tracker is not specifically optimized and can run faster with fewer parameters to get comparable results. While the pixel tracker model greatly outperforms instance-tracker methods. The HybridTracker achieves the best results owing to the fusion module. Fig.~\ref{fig:comparison ablation study hybridtracker temporal constraints} shows the qualitative results of our methods on Cityscapes-VPS. The qualitative results on the VIPER dataset shown in Fig.~\ref{fig:ablation study of Different Trackers viper} demonstrate that HybridTracker is expert in settling small objects with large motion (\whiteding{1} and \whiteding{2}). In addition, the temporal consistency constraints help track small and large objects with occlusion and inter-frame jump due to deformation (\whiteding{3}).


\subsubsection{\textbf{Analysis on Differentiable Matching Layer in Instance Tracker}}
To verify the effect of this layer, we explore the performance of instance embedding with various loss functions. Experiments show that our method can be trained more smoothly and can converge faster than the existing contrast loss with higher VPQ scores, shown in Fig.~\ref{fig:method.Training of Instance Tracker}.

\subsubsection{\textbf{Analysis on RAFT Version and Iterations}} In Pixel Tracker, we calculate the dice coefficient of different instances from the frame $t-\delta$ and frame $t$ given the optical flow map to form the position-aware similarity matrix. We use the flow estimation network RAFT to estimate the optical flow from two consecutive frames. The source code of RAFT is released with five versions of pre-trained weights. We performed ablation studies on different versions of RAFT to understand the impact on the pixel tracker, shown in Table~\ref{tab:city_vpq sensitivity pixel track}. The sintel version of RAFT achieves the best performance, for MIP Sintel is a large synthetic dataset that provides detailed and dense optical flow annotation, and it contains many challenging scenes, making it a good generalization capability. We perform different iterations on the raft-sintel to understand the impact on the pixel tracker, shown in Tab.~\ref{tab:city_vpq iteration raft pixel track}. RAFT with 20 iterations achieves the best performance. 

\begin{table}[t]
\caption{\textbf{Ablation study of different versions of RAFT, which affects pixel tracker on Cityscapes-VPS validation dataset.} The best results are highlighted in boldface.}
\centering
\resizebox{0.7\linewidth}{!}{%
\begin{tabular}[b]{ l|ccc}
        \hline
        version      & PQ & PQ\textsuperscript{Th}  & PQ\textsuperscript{St} \\
        \hline
        raft-chairs              & 57.01 & 44.08 & 66.42 \\
        raft-kitti  & 56.97 & 43.98 & 66.42  \\ 
 raft-sintel & \textbf{57.39} & \textbf{44.97} & \textbf{66.42}  \\ 
 raft-smalll & 57.09 & 44.26 & 66.42  \\ 
 raft-things & 57.35 & 44.87 & 66.42  \\

        \hline
\end{tabular}
}

\label{tab:city_vpq sensitivity pixel track}
\end{table}


\begin{table}[t]
\centering
\caption{\textbf{Ablation study of different iterations on sintel version of RAFT, which affects Pixel Tracker on Cityscapes-VPS val dataset.} The best results are highlighted in boldface.}
\resizebox{0.7\linewidth}{!}{%
\begin{tabular}[b]{ l|cccc}
        \hline
        Iterations      & PQ & PQ\textsuperscript{Th}  & PQ\textsuperscript{St} & FPS \\
        \hline
    1 & 57.74 & 45.81 & 66.42 & 4.06\\
    5  & 58.18 & 46.85 & 66.42 & 3.33 \\ 
    10 & 58.26 & 47.03 & 66.42 & 2.21\\ 
    15 & 58.27 & 47.06 & 66.42  & 1.52\\ 
    20 & \textbf{58.37} & \textbf{47.31} & \textbf{66.42} & 1.21  \\ 
    25 & 58.29 & 47.11 & 66.42 & 0.95 \\

        \hline
\end{tabular}
}

\label{tab:city_vpq iteration raft pixel track}
\end{table}



\begin{figure*}[!htbp]
\centering
\subfigure[Instance Tracker.]{
  \begin{minipage}[]{0.23\linewidth}
        \centering
        \includegraphics[width=4.45cm]{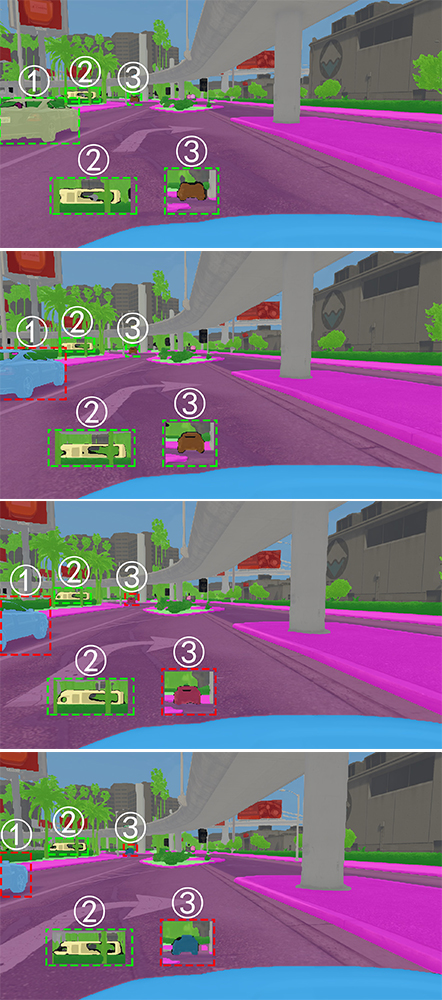}\\
    \end{minipage}%
    \label{fig:comparison Instance Tracker on viper}
}
    \subfigure[Pixel Tracker.]{
  \begin{minipage}[]{0.23\linewidth}
        \centering
        \includegraphics[width=4.45cm]{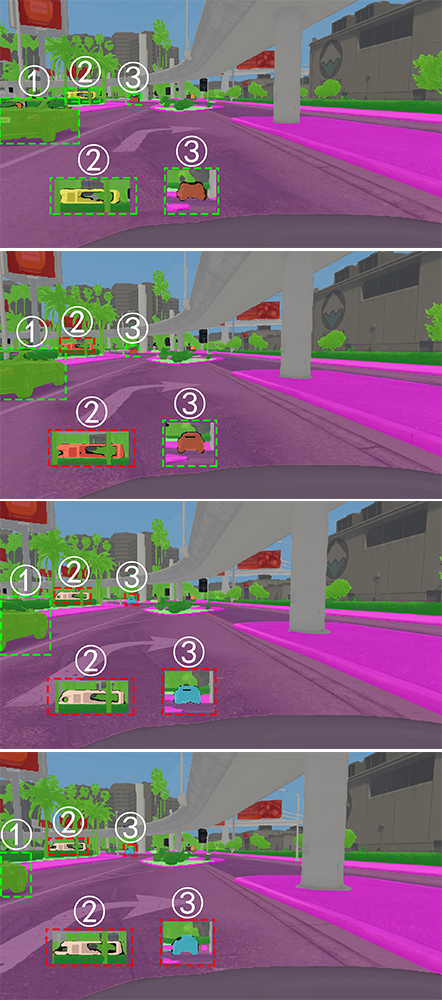}\\
    \end{minipage}%
    \label{fig:comparison val Pixel Tracker on viper}
  }
    \subfigure[HybridTracker.]{
  \begin{minipage}[]{0.23\linewidth}
        \centering
        \includegraphics[width=4.45cm]{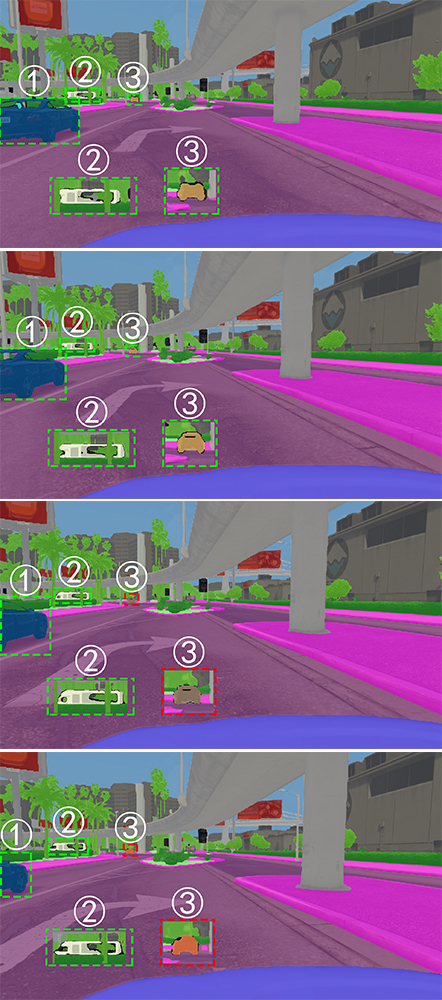}\\
    \end{minipage}%
    \label{fig:comparison Hybrid Tracker on viper}
  }
    \subfigure[HybridTracker + w/ Temporal.]{
  \begin{minipage}[]{0.23\linewidth}
        \centering
        \includegraphics[width=4.45cm]{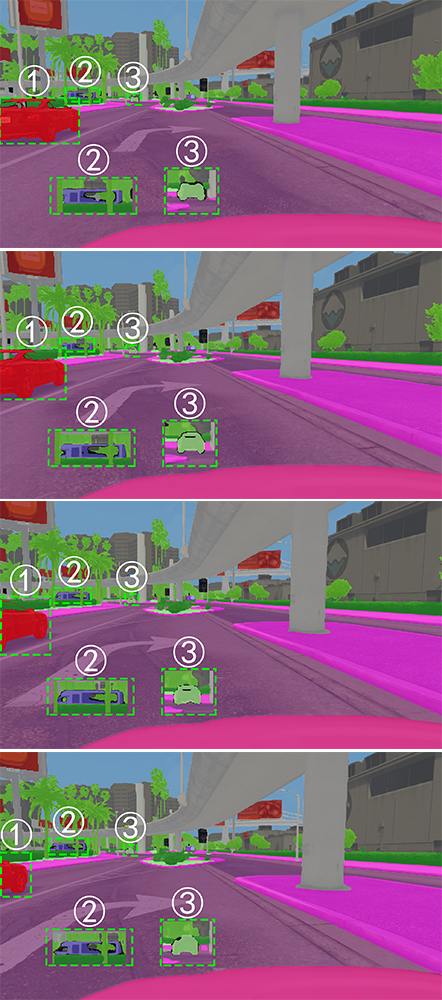}\\
    \end{minipage}%
    \label{fig:comparison Hybrid Tracker w/ Temporal on viper}
  }
  \caption{\textbf{Illustration of our different trackers on VIPER dataset.} Instance Tracker can handle large objects (with minor contour deformation but struggles to handle small objects (\whiteding{2}). In contrast, Pixel Tracker can handle small objects with limited motion, rather than irregular motion (\whiteding{1}). HybridTracker is expert in settling small objects with large motion (\whiteding{1} and \whiteding{2}). The temporal consistency constraints help track small and large objects with occlusion and inter-frame jump due to deformation (\whiteding{3}).}
\label{fig:ablation study of  Different Trackers viper}
\end{figure*}

\begin{figure}[t]
\centering
\subfigure[Instance Tracker.]{
  \begin{minipage}[]{\linewidth}
        \centering
        \includegraphics[width=0.99\linewidth]{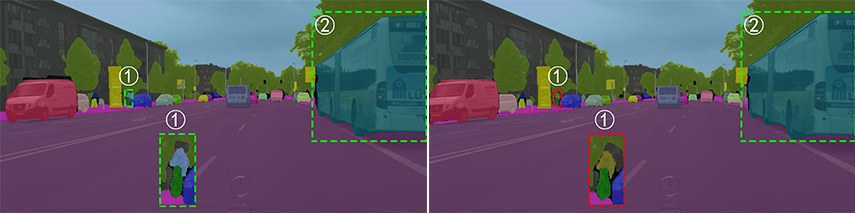}\\
        \vspace{0.10cm}
    \end{minipage}%
    \label{fig:illustration hybrid tracker}
  }
  
  \subfigure[Pixel Tracker.]{
  \begin{minipage}[]{\linewidth}
        \centering
        \includegraphics[width=0.99\linewidth]{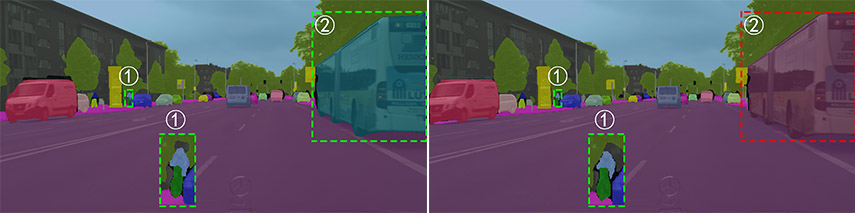}\\
        \vspace{0.10cm}
    \end{minipage}%
    \label{fig:illustration instance tracker with temporal}
  }
  \subfigure[HybridTracker.]{
  \begin{minipage}[]{\linewidth}
        \centering
        \includegraphics[width=0.99\linewidth]{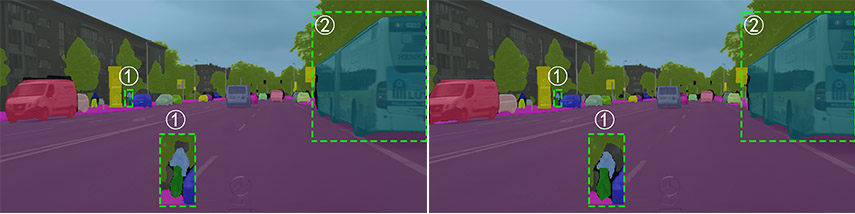}\\
        \vspace{0.10cm}
    \end{minipage}%
    \label{fig:illustration instance tracker with temporal}
  }
\caption{\textbf{Ablation study of different trackers on Cityscapes-VPS validation dataset.} The same color of instance masks means the same instance. The green dashed boxes indicate the instances are correctly matched, while the red dashed boxes indicate false matches. Instance Tracker can handle large objects with more minor contour deformations (\whiteding{2}) but struggles to handle small objects. In contrast, the pixel tracker can handle small objects with limited motions (\whiteding{1}). HybridTracker can robustly settle these two cases (\whiteding{1} and \whiteding{2}).}
\label{fig:ablation study of  Different Trackers}
\end{figure}



\begin{table*}[]
\caption{\textbf{Ablaton study of temporal consistency constraints on Cityscapes-VPS validation set with our method variants}. Each cell contains VPQ / VPQ\textsuperscript{Th} / VPQ\textsuperscript{St} scores. The best results are highlighted in boldface. Note that the configuration of our method is consistent with VPSNet-Track. All trackers with temporal consistency constraints have improved performance. The combination of VPSNet-FuseTrack and our pixel tracker with temporal consistency constraint outperforms VPSNet-FuseTrack. Our HybridTracker with temporal consistency constraint achieves the best VPQ scores.}
\centering
\resizebox{\textwidth}{!}{%
\begin{adjustbox}{max width=\textwidth}
\begin{tabular}{l|c|c|c|c| c|c}
\hline
{Methods} &\multicolumn{4}{c|}{Temporal window size} 
                & \multirow{2}{*}{VPQ} & \multirow{2}{*}{FPS}\\
\cline{2-5} {on \textbf{Cityscapes-VPS \textit{val.}}} & k = 0 & k = 5 & k = 10 & k = 15 &  \\
\hline
VPSNet-Track  & 
			63.1 / 56.4 / 68.0 & 
            56.1 / 44.1 / 64.9 & 
            53.1 / 39.0 / 63.4 &
            51.3 / 35.4 / 62.9 &   
            55.9 / 43.7 / 64.8 &
            4.5\\

VPSNet-FuseTrack \quad \quad \quad   & 
            64.5 / 58.1 / 69.1 & 
            57.4 / 45.2 / 66.4 &
            54.1 / 39.5 / 64.7 &   
            52.2 / 36.0 / 64.0 & 
            57.2 / 44.7 / 66.6 &
            1.3\\

SiamTrack &
64.6 / 58.3 / 69.1 &
57.6 / 45.6 / 66.6 &
54.2 / 39.2 / 65.2 &
52.7 / 36.7 / 64.6 &
57.3 / 44.7 / 66.4 &
4.5\\
            \hline
Instance Tracker  & 
			65.6 / 60.0 / 69.7 & 
            55.1 / 39.3 / 66.6 & 
            51.4 / 32.5 / 65.1 &
            49.3 / 28.6 / 64.3 & 
            55.3 / 40.1 / 66.4 &
            5.1 \\

Instance Tracker + w/ Mutual Check  & 
			65.6 / 60.0 / 69.7 & 
            55.0 / 39.2 / 66.6 & 
            51.5 / 32.7 / 65.1 &
            49.4 / 28.9 / 64.3 &   
            55.37 / 40.18 / 66.42 &
            5.0 \\

Instance Tracker + w/ Temporal  & 
			65.3 / 59.3 / 69.7 & 
            55.2 / 39.6 / 66.6 & 
            51.6 / 33.0 / 65.1 &
            49.4 / 29.0 / 64.3 &   
            55.38 / 40.21 / 66.42 &
            4.9 \\

Instance Tracker + w/ Mutual + Temporal  & 
			65.3 / 59.4 / 69.7 & 
            55.2 / 39.5 / 66.6 & 
            51.7 / 33.2 / 65.1 &
            49.6 / 29.3 / 64.3 &   
            55.44 / 40.35 / 66.42 & 
            4.8 \\
            
Pixel Tracker  & 
			65.6 / 60.0 / 69.7 & 
            58.8 / 48.0 / 66.6 & 
            55.4 / 42.1 / 65.1 &
            53.2 / 38.0 / 64.3 & 
            58.24 / 47.0 / 66.42 &
            3.8 \\
            
Pixel Track + w/ Temporal & 
			65.6 / 60.0 / 69.7 & 
            58.9 / 48.3 / 66.6 & 
            55.5 / 42.4 / 65.1 &
            53.3 / 38.2 / 64.3 & 
            58.34 / 47.1 / 66.42 &
            3.6 \\
            
HybridTracker (Instance + Pixel) & 
			65.6 / 60.0 / 69.7 & 
            58.9 / 48.3 / 66.6 & 
            55.6 / 42.5 / 65.1 &
            53.4 / 38.5 / 64.3 &  
            58.37 / 47.3 / 66.4 &
            3.6 \\
            
HybridTracker + w/ Temporal \quad \quad \quad   & 
            \textbf{65.6} / 60.0 / 69.7 & 
            \textbf{59.0} / 48.5 / 66.6 &
            \textbf{55.7} / 42.8 / 65.1 &   
            \textbf{53.6} / 38.9 / 64.3 & 
            \textbf{58.47} / 47.5 / 66.4 &
            3.3 \\
\hline

VPSNet-FuseTrack + Pixel  & 
			65.0 / 58.9 / 69.4 & 
            57.7 / 45.4 / 66.7 & 
            54.7 / 39.9 / 65.6 &
            53.1 / 36.5 / 65.3 &     
            57.6515 / 45.1600 /  66.7363 &
            1.1 \\
            
VPSNet-FuseTrack + Pixel + w/ Temporal  & 
			65.0 / 58.9 / 69.4 & 
            57.7 / 45.4 / 66.7 & 
            54.8 / 39.9 / 65.6 &
            53.1 / 36.5 / 65.3 &     
            57.6524 / 45.1619 / 66.7463 &
            0.9 \\
\hline
\end{tabular}
\end{adjustbox}
}

\label{tab:cityvps_vpq appendix cityscapes val fusetrack}
\end{table*}       

\begin{table*}[]
\caption{\textbf{Ablaton study of temporal consistency constraints on VIPER dataset with our method variants}. Each cell contains VPQ / VPQ\textsuperscript{Th} / VPQ\textsuperscript{St} scores. The best results are highlighted in boldface. \textbf{Note} that the configuration of our method is consistent with VPSNet-Track.  All trackers with temporal consistency constraints have improved performance. Our HybridTracker with temporal consistency constraint achieves the best VPQ scores.}
\centering
\resizebox{\textwidth}{!}{%
\begin{adjustbox}{max width=\textwidth}
\begin{tabular}{l|c|c|c|c| c|c}
\hline
{Methods} &\multicolumn{4}{c|}{Temporal window size} 
                & \multirow{2}{*}{VPQ} & \multirow{2}{*}{FPS}\\
 \cline{2-5} {on \textbf{VIPER}} & k = 0 & k = 5 & k = 10 & k = 15 &  \\
\hline
VPSNet-Track  & 
		    48.1 / 38.0 / 57.1 & 
            49.3 / 45.6 / 53.7 & 
            45.9 / 37.9 / 52.7 &
            43.2 / 33.6 / 51.6 &  
            46.6 / 38.8 / 53.8 &
            5.1 \\
            
VPSNet-FuseTrack      & 
			49.8 / 40.3 / 57.7 & 
            51.6 / 49.0 / 53.8 & 
            47.2 / 40.4 / 52.8 &
            45.1 / 36.5 / 52.3 &   
            48.4 / 41.6 / 53.2 &
            1.6 \\
SiamTrack &
            51.1 / 42.3 / 58.5 &
            \textbf{53.4} / 51.9 / 54.6 &
            49.2 / 44.1 / 53.5 &
            47.2 / 40.3 / 52.9 &
            50.2 / 44.7 / 55.0 &
            5.1 \\
\hline
Instance Tracker  & 
			55.1 / 51.2 / 58.0 & 
            45.7 / 30.4 / 57.5 & 
            43.8 / 26.2 / 57.3 &
            41.5 / 21.5 / 57.0 &     
            46.53 / 32.33 / 57.45 & 
            5.7 \\
            
Instance Tracker + w/ Temporal & 
			54.4 / 49.7 / 58.0 & 
            46.1 / 31.3 / 57.5 & 
            44.4 / 27.6 / 57.3 &
            42.3 / 23.2 / 57.0 &     
            46.79 / 32.95 / 57.45 &
            5.5 \\
            
Pixel Tracker  & 
			\textbf{55.1} / 51.2 / 58.0 & 
            52.2 / 45.4 / 57.5 & 
            50.9 / 42.4 / 57.3 &
            49.5 / 39.9 / 57.0 &     
            51.92 / 44.73 / 57.45 &
            3.8 \\

Pixel Tracker + w/ Temporal   & 
			55.0 / 51.2 / 58.0 & 
            52.5 / 46.0 / 57.5 & 
            51.3 / 43.4 / 57.3 &
            50.2 / 41.3 / 57.0 &     
            52.24 / 45.48 / 57.45 &
            3.7 \\

HybridTracker  & 
			55.0 / 51.2 / 58.0 & 
            52.4 / 45.7 / 57.5 & 
            51.0 / 42.8 / 57.3 &
            49.8 / 40.4 / 57.0 &     
            52.06 / 45.05 / 57.45 &
            3.6 \\

HybridTracker + w/ Temporal  & 
			55.0 / 51.2 / 58.0 & 
            52.6 / 46.2 / 57.5 & 
            \textbf{51.5} / 43.9 / 57.3 &
            \textbf{50.4} / 41.9 / 57.0 &     
            \textbf{52.39} / 45.80 / 57.45 &
            3.5 \\
            
\hline
\end{tabular}
\end{adjustbox}
}

\label{tab:viper_vpq appendix temporal viper}
\end{table*}

\begin{table*}[tbh!]
\caption{\textbf{Ablation study of the different trackers under different temporal thresholds on Cityscapes-VPS validation dataset.} Each cell contains VPQ / VPQ\textsuperscript{Th} / VPQ\textsuperscript{St} scores. The best results are highlighted in boldface. The change of the thresholds of temporal is not sensitive to VPQs. }
\centering
\resizebox{\textwidth}{!}{%
\begin{adjustbox}{max width=\textwidth}
\begin{tabular}{l|c|c|c|c|c}
\hline
{Methods} &\multicolumn{5}{c}{Temporal Thresholds $\theta$} \\
\cline{2-6} {on \textbf{Cityscapes-VPS \textit{val.}}} & $\theta$ = 0.001 & $\theta$ = 0.005 & $\theta$ = 0.01 & $\theta$ = 0.015 & $\theta$ = 0.02 \\

\hline

\hline
Instance Tracker & 
                55.435/40.331/66.420 & 
                \textbf{55.445}/40.355/66.420 & 
                55.444/40.353/66.420 &
                55.441/40.344/66.420 &
                55.440/40.342/66.420 \\
                
 Pixel Tracker  & 
                57.467/45.157/66.420 &
                57.467/45.157/66.420 &
                \textbf{57.468}/45.160/66.420 &
                57.468/45.160/66.420 &
                57.468/45.160/66.420 \\
                
 HybridTracker & 
                57.564/45.387/66.420 &
                57.565/45.389/66.420 &
                \textbf{57.568}/45.398/66.420 &
                57.568/45.398/66.420 &
                57.568/45.398/66.420 \\

\hline
\end{tabular}
\end{adjustbox}
}
\\


\label{tab:cityvps_vpq val sensitivity temporal}
\end{table*}

\begin{figure}[hbtp]
\centering
\subfigure[HybridTracker.]{
  \begin{minipage}{4.2cm}
        \centering
        \includegraphics[width=4.2cm]{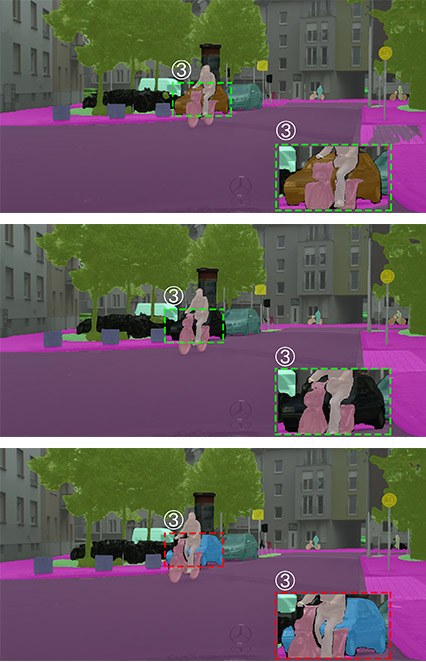}\\
    \end{minipage}%
  }
  \subfigure[HybridTracker + w/ Temporal.]{
  \begin{minipage}{4.2cm}
        \centering
        \includegraphics[width=4.2cm]{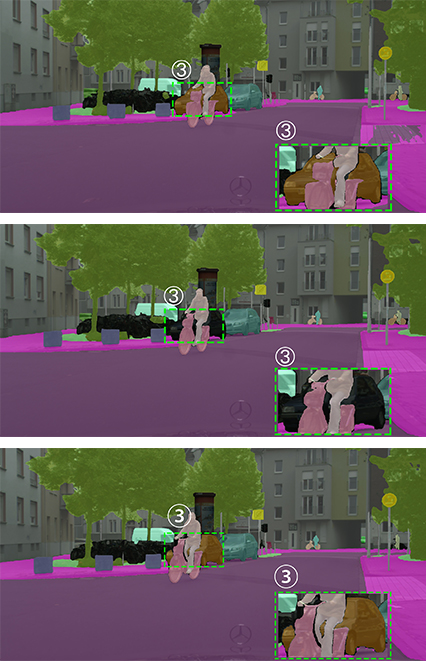}\\
    \end{minipage}%
  }
\caption{\textbf{HybridTrack w/o vs w/ temporal consistency constraint on Cityscapes-VPS.} The car in the second image cannot be segmented due to occlusion, resulting in a new id, while the temporal consistency constraint can circumvent this problem (\whiteding{3}).}
\label{fig:Hybrid Track w/o vs w/ Temporal Consistency Constraint on cityscapes}
\end{figure}



\subsubsection{\textbf{Analysis on Mutual Check and Temporal Consistency Constraints}}
\label{sec:appendix.Ablation Study of Mutual Check and Temporal Consistency Constraints}
We also conducted additional experiments on the mutual check and temporal consistency constraints with ablation consideration. All of our methods (including Instance Tracker, Pixel Tracker, and HybridTracker) with mutual check and temporal consistency constraints achieve different levels of improvement of VPQs, illustrated in Tab.~\ref{tab:cityvps_vpq appendix cityscapes val fusetrack} and ~\ref{tab:viper_vpq appendix temporal viper}. Fig.~\ref{fig:ablation study of  Different Trackers viper} and ~\ref{fig:Hybrid Track w/o vs w/ Temporal Consistency Constraint on cityscapes} demonstrate the improvements of the temporal consistency constraint. In addition, we combine better panoptic segmentation results from VPSNet-FuseTrack~\cite{kim2020vps} with our Pixel Tracker on the Cityscpaes-VPS validation dataset to achieve better VPQ scores. Our HybridTracker with temporal consistency constraint achieves the best performance on Cityscapes-VPS and VIPER validation datasets, shown in Tab.~\ref{tab:cityvps_vpq appendix cityscapes val fusetrack} and ~\ref{tab:viper_vpq appendix temporal viper}.

\subsubsection{\textbf{Analysis on Temporal Thresholds}} We propose mutual check and temporal consistency constraints in inference to effectively address complex challenges, such as occlusion and strong contour deformations. We randomly select 3 images to form the image set, respectively: frame $t-\delta$, frame $t$, and frame $t+\delta$. To strengthen the matching relationship between adjacent frames and reduce false matches across frames, the temporal consistency constraint requires that: only when the similarity values of frame $t-\delta$ and frame $t+\delta$ are greater than a certain threshold, a cross-frame matching relationship will be established, assigning the ID of instance $i$ in frame $t-\delta$ to the ID of instance $j$ in frame $t+\delta$. We set the thresholds to 0.001, 0.005, 0.01, 0.015 and 0.02. As shown in Tab.~\ref{tab:cityvps_vpq val sensitivity temporal}, the temporal consistency thresholds do not have a great impact on VPQ results.










\section{Conclusion}
We propose a lightweight and efficient hybrid tracker network consisting of an instance tracker and pixel tracker that effectively copes with video panoptic segmentation. Our proposed differentiable matching layer makes the instance tracker training stable and converges fast, while the dense pixel tracker can settle the small objects with smooth motion. The mutual check and temporal consistency constraints can effectively address complex challenges, such as occlusion and strong contour deformations. Comprehensive experiments show that HybridTracker achieves superior performance than state-of-the-art methods on Cityscapes-VPS and VIPER datasets.

\noindent\textbf{Limitations.} At first, the speed and accuracy of the tracking in our method are restricted to the optical flow estimation, while we find that RAFT~\cite{teed2020raft} performs well on all datasets. The performance and efficiency of our method can be further improved with more advanced optical flow estimation methods.
 Second, our work focuses only on improving the tracking of individual instances rather than fusion for better segmentation results. Thus the mutual promotion of tracking and segmentation is a worthwhile direction to investigate.



\bibliographystyle{IEEEtran}
\bibliography{HybridTracker}

\begin{thebibliography}{10}
\providecommand{\url}[1]{#1}
\csname url@samestyle\endcsname
\providecommand{\newblock}{\relax}
\providecommand{\bibinfo}[2]{#2}
\providecommand{\BIBentrySTDinterwordspacing}{\spaceskip=0pt\relax}
\providecommand{\BIBentryALTinterwordstretchfactor}{4}
\providecommand{\BIBentryALTinterwordspacing}{\spaceskip=\fontdimen2\font plus
\BIBentryALTinterwordstretchfactor\fontdimen3\font minus
  \fontdimen4\font\relax}
\providecommand{\BIBforeignlanguage}[2]{{%
\expandafter\ifx\csname l@#1\endcsname\relax
\typeout{** WARNING: IEEEtran.bst: No hyphenation pattern has been}%
\typeout{** loaded for the language `#1'. Using the pattern for}%
\typeout{** the default language instead.}%
\else
\language=\csname l@#1\endcsname
\fi
#2}}
\providecommand{\BIBdecl}{\relax}
\BIBdecl

\bibitem{kim2020vps}
D.~Kim, S.~Woo, J.-Y. Lee, and I.~S. Kweon, ``{Video Panoptic Segmentation},''
  in \emph{Proceedings of the IEEE Conference on Computer Vision and Pattern
  Recognition (CVPR)}, 2020.

\bibitem{xiong2019upsnet}
Y.~Xiong, R.~Liao, H.~Zhao, R.~Hu, M.~Bai, E.~Yumer, and R.~Urtasun, ``{UPSNet:
  A Unified Panoptic Segmentation Network},'' in \emph{Proceedings of the
  IEEE/CVF Conference on Computer Vision and Pattern Recognition (CVPR)}, 2019,
  pp. 8818--8826.

\bibitem{yang2019video}
L.~Yang, Y.~Fan, and N.~Xu, ``{Video Instance Segmentation},'' in
  \emph{Proceedings of the IEEE/CVF International Conference on Computer Vision
  (ICCV)}, 2019, pp. 5188--5197.

\bibitem{woo2021learning}
S.~Woo, D.~Kim, J.-Y. Lee, and I.~S. Kweon, ``{Learning to Associate Every
  Segment for Video Panoptic Segmentation},'' in \emph{Proceedings of the
  IEEE/CVF Conference on Computer Vision and Pattern Recognition (CVPR)}, 2021,
  pp. 2705--2714.

\bibitem{zhou2020tracking}
X.~Zhou, V.~Koltun, and P.~Kr{\"a}henb{\"u}hl, ``{Tracking Objects as
  Points},'' in \emph{Proceedings of the European Conference on Computer Vision
  (ECCV)}.\hskip 1em plus 0.5em minus 0.4em\relax Springer, 2020, pp. 474--490.

\bibitem{li2021panopticfcn}
Y.~Li, H.~Zhao, X.~Qi, L.~Wang, Z.~Li, J.~Sun, and J.~Jia, ``{Fully
  Convolutional Networks for Panoptic Segmentation},'' in \emph{Proceedings of
  the IEEE Conference on Computer Vision and Pattern Recognition (CVPR)}, 2021,
  pp. 214--223.

\bibitem{hermans2017defense}
A.~Hermans, L.~Beyer, and B.~Leibe, ``{In Defense of the Triplet Loss for
  Person Re-Identification},'' \emph{arXiv preprint arXiv:1703.07737}, 2017.

\bibitem{Ye2021SuperPlane}
W.~Ye, H.~Li, T.~Zhang, X.~Zhou, H.~Bao, and G.~Zhang, ``{SuperPlane: 3D Plane
  Detection and Description from a Single Image},'' in \emph{2021 IEEE Virtual
  Reality and 3D User Interfaces (VR)}, 2021, pp. 207--215.

\bibitem{teed2020raft}
Z.~Teed and J.~Deng, ``{RAFT: Recurrent All-Pairs Field Transforms for Optical
  Flow},'' in \emph{Proceedings of the European Conference on Computer Vision
  (ECCV)}.\hskip 1em plus 0.5em minus 0.4em\relax Springer, 2020, pp. 402--419.

\bibitem{wang2021survey}
W.~Wang, T.~Zhou, F.~Porikli, D.~Crandall, and L.~Van~Gool, ``{A Survey on Deep
  Learning Technique for Video Segmentation},'' \emph{arXiv preprint
  arXiv:2107.01153}, 2021.

\bibitem{li2021video}
J.~Li, W.~Wang, J.~Chen, L.~Niu, J.~Si, C.~Qian, and L.~Zhang, ``{Video
  Semantic Segmentation via Sparse Temporal Transformer},'' in
  \emph{Proceedings of the 29th ACM International Conference on Multimedia (ACM
  MM)}, 2021, pp. 59--68.

\bibitem{hoyer2021three}
L.~Hoyer, D.~Dai, Y.~Chen, A.~Koring, S.~Saha, and L.~Van~Gool, ``{Three Ways
  to Improve Semantic Segmentation with Self-Supervised Depth Estimation},'' in
  \emph{Proceedings of the IEEE/CVF Conference on Computer Vision and Pattern
  Recognition (CVPR)}, 2021, pp. 11\,130--11\,140.

\bibitem{chen2020naive}
L.-C. Chen, R.~G. Lopes, B.~Cheng, M.~D. Collins, E.~D. Cubuk, B.~Zoph,
  H.~Adam, and J.~Shlens, ``{Naive-Student: Leveraging Semi-Supervised Learning
  in Video Sequences for Urban Scene Segmentation},'' in \emph{Proceedings of
  the European Conference on Computer Vision (ECCV)}.\hskip 1em plus 0.5em
  minus 0.4em\relax Springer, 2020, pp. 695--714.

\bibitem{liu2020efficient}
Y.~Liu, C.~Shen, C.~Yu, and J.~Wang, ``{Efficient Semantic Video Segmentation
  with Per-frame Inference},'' in \emph{Proceedings of the European Conference
  on Computer Vision (ECCV)}.\hskip 1em plus 0.5em minus 0.4em\relax Springer,
  2020, pp. 352--368.

\bibitem{yin2021learning}
Z.~Yin, J.~Zheng, W.~Luo, S.~Qian, H.~Zhang, and S.~Gao, ``{Learning to
  Recommend Frame for Interactive Video Object Segmentation in the Wild},'' in
  \emph{Proceedings of the IEEE/CVF Conference on Computer Vision and Pattern
  Recognition (CVPR)}, 2021, pp. 15\,445--15\,454.

\bibitem{heo2021guided}
Y.~Heo, Y.~J. Koh, and C.-S. Kim, ``{Guided Interactive Video Object
  Segmentation Using Reliability-Based Attention Maps},'' in \emph{Proceedings
  of the IEEE/CVF Conference on Computer Vision and Pattern Recognition
  (CVPR)}, 2021, pp. 7322--7330.

\bibitem{cheng2021modular}
H.~K. Cheng, Y.-W. Tai, and C.-K. Tang, ``{Modular Interactive Video Object
  Segmentation: Interaction-to-Mask, Propagation and Difference-Aware
  Fusion},'' in \emph{Proceedings of the IEEE/CVF Conference on Computer Vision
  and Pattern Recognition (CVPR)}, 2021, pp. 5559--5568.

\bibitem{xu2020segment}
Z.~Xu, W.~Zhang, X.~Tan, W.~Yang, H.~Huang, S.~Wen, E.~Ding, and L.~Huang,
  ``{Segment as Points for Efficient Online Multi-Object Tracking and
  Segmentation},'' in \emph{Proceedings of the European Conference on Computer
  Vision (ECCV)}.\hskip 1em plus 0.5em minus 0.4em\relax Springer, 2020, pp.
  264--281.

\bibitem{li2020delving}
Y.~Li, N.~Xu, J.~Peng, J.~See, and W.~Lin, ``{Delving into the Cyclic Mechanism
  in Semi-supervised Video Object Segmentation},'' \emph{Advances in Neural
  Information Processing Systems (NeurIPS)}, vol.~33, pp. 1218--1228, 2020.

\bibitem{xie2021efficient}
H.~Xie, H.~Yao, S.~Zhou, S.~Zhang, and W.~Sun, ``{Efficient Regional Memory
  Network for Video Object Segmentation},'' in \emph{Proceedings of the
  IEEE/CVF Conference on Computer Vision and Pattern Recognition (CVPR)}, 2021,
  pp. 1286--1295.

\bibitem{duke2021sstvos}
B.~Duke, A.~Ahmed, C.~Wolf, P.~Aarabi, and G.~W. Taylor, ``{SSTVOS: Sparse
  Spatiotemporal Transformers for Video Object Segmentation},'' in
  \emph{Proceedings of the IEEE/CVF Conference on Computer Vision and Pattern
  Recognition (CVPR)}, 2021, pp. 5912--5921.

\bibitem{ren2021reciprocal}
S.~Ren, W.~Liu, Y.~Liu, H.~Chen, G.~Han, and S.~He, ``{Reciprocal
  Transformations for Unsupervised Video Object Segmentation},'' in
  \emph{Proceedings of the IEEE/CVF Conference on Computer Vision and Pattern
  Recognition (CVPR)}, 2021, pp. 15\,455--15\,464.

\bibitem{porzi2020learning}
L.~Porzi, M.~Hofinger, I.~Ruiz, J.~Serrat, S.~R. Bulo, and P.~Kontschieder,
  ``{Learning Multi-Object Tracking and Segmentation from Automatic
  Annotations},'' in \emph{Proceedings of the IEEE/CVF Conference on Computer
  Vision and Pattern Recognition (CVPR)}, 2020, pp. 6846--6855.

\bibitem{luiten2020unovost}
J.~Luiten, I.~E. Zulfikar, and B.~Leibe, ``{UnOVOST: Unsupervised Offline Video
  Object Segmentation and Tracking},'' in \emph{Proceedings of the IEEE/CVF
  Winter Conference on Applications of Computer Vision (WACV)}, 2020, pp.
  2000--2009.

\bibitem{yang2021dystab}
Y.~Yang, B.~Lai, and S.~Soatto, ``{DyStaB: Unsupervised Object Segmentation via
  Dynamic-Static Bootstrapping},'' in \emph{Proceedings of the IEEE/CVF
  Conference on Computer Vision and Pattern Recognition (CVPR)}, 2021, pp.
  2826--2836.

\bibitem{hui2021collaborative}
T.~Hui, S.~Huang, S.~Liu, Z.~Ding, G.~Li, W.~Wang, J.~Han, and F.~Wang,
  ``{Collaborative Spatial-Temporal Modeling for Language-Queried Video Actor
  Segmentation},'' in \emph{Proceedings of the IEEE/CVF Conference on Computer
  Vision and Pattern Recognition (CVPR)}, 2021, pp. 4187--4196.

\bibitem{ye2021referring}
L.~Ye, M.~Rochan, Z.~Liu, X.~Zhang, and Y.~Wang, ``{Referring Segmentation in
  Images and Videos with Cross-Modal Self-Attention Network},'' \emph{arXiv
  preprint arXiv:2102.04762}, 2021.

\bibitem{mcintosh2020visual}
B.~McIntosh, K.~Duarte, Y.~S. Rawat, and M.~Shah, ``{Visual-Textual Capsule
  Routing for Text-Based Video Segmentation},'' in \emph{Proceedings of the
  IEEE/CVF Conference on Computer Vision and Pattern Recognition (CVPR)}, 2020,
  pp. 9942--9951.

\bibitem{zhou2021target}
T.~Zhou, J.~Li, X.~Li, and L.~Shao, ``Target-aware object discovery and
  association for unsupervised video multi-object segmentation,'' in
  \emph{Proceedings of the IEEE/CVF Conference on Computer Vision and Pattern
  Recognition (CVPR)}, 2021, pp. 6985--6994.

\bibitem{ventura2019rvos}
C.~Ventura, M.~Bellver, A.~Girbau, A.~Salvador, F.~Marques, and X.~Giro-i
  Nieto, ``{RVOS: End-to-End Recurrent Network for Video Object
  Segmentation},'' in \emph{Proceedings of the IEEE/CVF Conference on Computer
  Vision and Pattern Recognition (CVPR)}, 2019, pp. 5277--5286.

\bibitem{wang2020end}
Y.~Wang, Z.~Xu, X.~Wang, C.~Shen, B.~Cheng, H.~Shen, and H.~Xia, ``{End-to-End
  Video Instance Segmentation with Transformers},'' in \emph{Proceedings of the
  IEEE/CVF Conference on Computer Vision and Pattern Recognition (CVPR)}, 2021.

\bibitem{athar2020stem}
A.~Athar, S.~Mahadevan, A.~Osep, L.~Leal-Taix{\'e}, and B.~Leibe, ``{STEm-Seg:
  Spatio-temporal Embeddings for Instance Segmentation in Videos},'' in
  \emph{Proceedings of the European Conference on Computer Vision
  (ECCV)}.\hskip 1em plus 0.5em minus 0.4em\relax Springer, 2020, pp. 158--177.

\bibitem{bertasius2020classifying}
G.~Bertasius and L.~Torresani, ``{Classifying, Segmenting, and Tracking Object
  Instances in Video with Mask Propagation},'' in \emph{Proceedings of the
  IEEE/CVF Conference on Computer Vision and Pattern Recognition (CVPR)}, 2020,
  pp. 9739--9748.

\bibitem{Cao_SipMask_ECCV_2020}
J.~Cao, R.~M. Anwer, H.~Cholakkal, F.~S. Khan, Y.~Pang, and L.~Shao,
  ``{SipMask: Spatial Information Preservation for Fast Image and Video
  Instance Segmentation},'' in \emph{Proceedings of the European Conference on
  Computer Vision (ECCV)}, 2020.

\bibitem{ying2021srnet}
X.~Ying, X.~Li, and M.~C. Chuah, ``{SRNet: Spatial Relation Network for
  Efficient Single-stage Instance Segmentation in Videos},'' in
  \emph{Proceedings of the 29th ACM International Conference on Multimedia (ACM
  MM)}, 2021, pp. 347--356.

\bibitem{lin2021video}
H.~Lin, R.~Wu, S.~Liu, J.~Lu, and J.~Jia, ``{Video Instance Segmentation with a
  Propose-Reduce Paradigm},'' in \emph{Proceedings of the IEEE/CVF
  International Conference on Computer Vision (ICCV)}, 2021, pp. 1739--1748.

\bibitem{qiao2021vip}
S.~Qiao, Y.~Zhu, H.~Adam, A.~Yuille, and L.-C. Chen, ``{ViP-DeepLab: Learning
  Visual Perception with Depth-aware Video Panoptic Segmentation},'' in
  \emph{Proceedings of the IEEE/CVF Conference on Computer Vision and Pattern
  Recognition (CVPR)}, 2021, pp. 3997--4008.

\bibitem{Weber2021NEURIPSDATA}
M.~Weber, J.~Xie, M.~Collins, Y.~Zhu, P.~Voigtlaender, H.~Adam, B.~Green,
  A.~Geiger, B.~Leibe, D.~Cremers, A.~Osep, L.~Leal-Taixe, and L.-C. Chen,
  ``{STEP: Segmenting and Tracking Every Pixel},'' in \emph{Neural Information
  Processing Systems (NeurIPS) Track on Datasets and Benchmarks}, 2021.

\bibitem{zhang2020fair}
Y.~Zhang, C.~Wang, X.~Wang, W.~Zeng, and W.~Liu, ``{FairMOT: On the Fairness of
  Detection and Re-Identification in Multiple Object Tracking},'' \emph{arXiv
  preprint arXiv:2004.01888}, 2020.

\bibitem{Pang_2021_CVPR}
J.~Pang, L.~Qiu, X.~Li, H.~Chen, Q.~Li, T.~Darrell, and F.~Yu, ``{Quasi-Dense
  Similarity Learning for Multiple Object Tracking},'' in \emph{Proceedings of
  the IEEE/CVF Conference on Computer Vision and Pattern Recognition (CVPR)},
  June 2021, pp. 164--173.

\bibitem{wang2019towards}
Z.~Wang, L.~Zheng, Y.~Liu, and S.~Wang, ``{Towards Real-Time Multi-Object
  Tracking},'' \emph{arXiv preprint arXiv:1909.12605}, 2019.

\bibitem{He_2017_ICCV}
K.~He, G.~Gkioxari, P.~Dollar, and R.~Girshick, ``{Mask R-CNN},'' in
  \emph{Proceedings of the IEEE International Conference on Computer Vision
  (ICCV)}, Oct 2017.

\bibitem{milletari2016v}
F.~Milletari, N.~Navab, and S.-A. Ahmadi, ``{V-Net: Fully Convolutional Neural
  Networks for Volumetric Medical Image Segmentation},'' in \emph{Proceedings
  of the International Conference on 3D Vision (3DV)}.\hskip 1em plus 0.5em
  minus 0.4em\relax IEEE, 2016, pp. 565--571.

\bibitem{lin2018focal}
T.-Y. Lin, P.~Goyal, R.~Girshick, K.~He, and P.~Dollár, ``{Focal Loss for
  Dense Object Detection},'' 2018.

\bibitem{yu2020learning}
H.~Yu, W.~Ye, Y.~Feng, H.~Bao, and G.~Zhang, ``{Learning Bipartite Graph
  Matching for Robust Visual Localization},'' in \emph{2020 IEEE International
  Symposium on Mixed and Augmented Reality (ISMAR)}.\hskip 1em plus 0.5em minus
  0.4em\relax IEEE, 2020, pp. 146--155.

\bibitem{wang2020learning}
Q.~Wang, X.~Zhou, B.~Hariharan, and N.~Snavely, ``{Learning Feature Descriptors
  using Camera Pose Supervision},'' in \emph{Proceedings of the European
  Conference on Computer Vision (ECCV)}, 2020.

\bibitem{CVPR2019_CycleTime}
X.~Wang, A.~Jabri, and A.~A. Efros, ``{Learning Correspondence from the
  Cycle-Consistency of Time},'' in \emph{Proceedings of the IEEE/CVF Conference
  on Computer Vision and Pattern Recognition (CVPR)}, 2019.

\end{thebibliography}

\end{document}